\documentclass{article}

    \PassOptionsToPackage{numbers}{natbib}

    \usepackage[final]{neurips_2024}

\usepackage[utf8]{inputenc} 
\usepackage[T1]{fontenc}    
\usepackage{hyperref}       
\usepackage{url}            
\usepackage{booktabs}       
\usepackage{amsfonts}       
\usepackage{nicefrac}       
\usepackage{microtype}      
\usepackage{xcolor}         
\usepackage{amsmath}
\usepackage{amssymb}
\usepackage{amsthm}
\usepackage{bbm}
\usepackage{algorithm}
\usepackage{algorithmic}
\usepackage{graphicx}
\hypersetup{      
                    colorlinks=true,                
                    citecolor=blue,
                    linkcolor=black,
}
\newcommand{\regret}{{\normalfont\textsc{Regret}}}

\title{Adversarial Environment Design via Regret-Guided Diffusion Models}

\author{%
  Hojun Chung$^1$, Junseo Lee$^1$, Minsoo Kim$^1$, Dohyeong Kim$^2$, and Songhwai Oh$^{1, 2,}$\thanks{Corresponding author: Songhwai Oh}\\
  \\
  $^1$Interdisciplinary Program in Artificial Intelligence and ASRI, Seoul National University\\
  $^2$Department of Electrical and Computer Engineering and ASRI, Seoul National University\\
  \texttt{\{hojun.chung, junseo.lee, minsoo.kim, dohyeong.kim\}@rllab.snu.ac.kr,}\\
  \texttt{songhwai@snu.ac.kr}
}

\newtheorem{theorem}{Theorem}[section]

\newtheorem{lemma}[theorem]{Lemma}
\newtheorem{proposition}[theorem]{Proposition}
\begin{document}

\maketitle

\begin{abstract}
Training agents that are robust to environmental changes remains a significant challenge in deep reinforcement learning (RL). Unsupervised environment design (UED) has recently emerged to address this issue by generating a set of training environments tailored to the agent's capabilities. While prior works demonstrate that UED has the potential to learn a robust policy, their performance is constrained by the capabilities of the environment generation. To this end, we propose a novel UED algorithm, adversarial environment design via regret-guided diffusion models (ADD). The proposed method guides the diffusion-based environment generator with the regret of the agent to produce environments that the agent finds challenging but conducive to further improvement. By exploiting the representation power of diffusion models, ADD can directly generate adversarial environments while maintaining the diversity of training environments, enabling the agent to effectively learn a robust policy. Our experimental results demonstrate that the proposed method successfully generates an instructive curriculum of environments, outperforming UED baselines in zero-shot generalization across novel, out-of-distribution environments. Project page: \href{https://rllab-snu.github.io/projects/ADD}{https://rllab-snu.github.io/projects/ADD}
\end{abstract}

\section{Introduction}
Deep reinforcement learning (RL) has achieved great success in various challenging domains, such as Atari \cite{atari}, GO \cite{go}, and real-world robotics tasks \cite{robot1, robot2}. Despite the progress, the deep RL agent struggles with the generalization problem; it often fails in unseen environments even with a small difference from the training environment distribution \cite{generalization1, generalization2}. To train well-generalizing policies, various prior works have used domain randomization (DR) \cite{dr_jakobi, dr_sadeghi, dr_tobin}, which provides RL agents with randomly generated environments. While DR enhances the diversity of the training environments, it requires a large number of trials to generate meaningful structures in high-dimensional domains. 
Curriculum reinforcement learning \cite{curriresearch1, curriresearch2} has been demonstrated to address these issues by providing instructive sequences of environments. Since manually designing an effective curriculum for complicated tasks is challenging, prior works \cite{poet, curripaper1} focus on generating curricula that consider the current agent's capabilities. Recently, unsupervised environment design (UED, \cite{paired}) has emerged as a scalable approach, notable for its advantage of requiring no prior knowledge. UED algorithms alternate between training the policy and designing training environments that maximize the regret of the agent. This closed-loop framework ensures the agent learns a minimax regret policy \cite{minimax_regret}, assuming that the two-player game between the agent and the environment generator reaches the Nash equilibrium.

There are two main approaches for UED: 1) learning-based methods, which employ an environment generator trained via reinforcement learning, and 2) replay-based methods, which selectively replay among previously generated environments. The learning-based methods \cite{paired, clutr, shed} utilize an adaptive generator that controls the parameters that fully define the environment configuration. The generator receives a regret of the agent as a reward and is trained via reinforcement learning to produce environments that maximize the regret. While the learning-based methods can directly generate meaningful environments, training the generator with RL is unstable due to the moving manifold \cite{clutr}. Additionally, we observe that the RL-based generator has limited environment coverage, which limits the generalization capability of the trained agent. In contrast, the replay-based methods \cite{plr, repaired, accel} employ a random generator and select environments to revisit among previously generated environments. Since the random generator can produce diverse environments without additional training, they outperform the learning-based methods in zero-shot generalization tasks \cite{accel}. However, the replay-based methods are sample inefficient as they require additional episodes to evaluate the regret on the randomly generated environments.

In this work, we propose a sample-efficient and robust UED algorithm by leveraging the strong representation power of diffusion models \cite{ddpm}. First, to make UED suitable for using a diffusion model as a generator, we introduce soft UED, which augments the regret objective of UED with an entropy regularization term, as done in maximum entropy RL \cite{maxent}. By incorporating the entropy term, we can ensure the diversity of the generated environments. Then, we present \textit{adversarial environment design via regret-guided diffusion models} (ADD), which guides a diffusion-based environment generator with the regret of the agent to produce environments that are conducive to the performance improvement of the agent. Enabling this regret guidance requires the gradient of the regret with respect to the environment parameter. However, since the true value of the regret is intractable and the regret estimation methods used in prior works on UED are not differentiable, a new form of regret estimation method is needed. To this end, we propose a novel method that enables the estimation of the regret in a differentiable form by utilizing an environment critic, which predicts a return distribution of the current policy on the given environment. This enables us to effectively integrate diffusion models within the UED framework, significantly enhancing the environment generation capability.


Since the regret-guided diffusion does not require an additional training of the environment generator, we can preserve the ability to cover the high-dimensional environment domain as the random generator of the replay-based method. Moreover, ADD can directly generate meaningful environments via regret-guided sampling as the learning-based methods. By doing so, ADD effectively combines the strengths of previous UED methods while addressing some of their limitations. Additionally, unlike other UED methods, ADD allows us to control the difficulty levels of the environments it generates by guiding the generator with the probability of achieving a specific return. It enables the reuse of the learned generator in various applications, such as generating benchmarks.

We conduct extensive experiments across challenging tasks commonly used in UED research: partially observable maze navigation and 2D bipedal locomotion over challenging terrain. Experimental results show that ADD achieves higher zero-shot generalization performance in unseen environments compared to the baselines. Furthermore, our analysis on the generated environments demonstrates that ADD produces an instructive curriculum with varying complexity while covering a large environment configuration space. As a result, it is shown that the proposed method successfully generates adversarial environments and facilitates the agent to learn a policy with solid generalization capabilities.

\section{Related Work}
\subsection{Unsupervsied Curriculum Reinforcement Learning}
While curriculum reinforcement learning \cite{curripaper1, curripaper2, curripaper3} has been shown to enhance the generalization performance of the RL agent, Dennis et al. \cite{paired} first introduce the concept of the unsupervised environment design (UED). UED encompasses various environment generation mehods, such as POET \cite{poet, poet2} and GPN\cite{from_nothing}. In this work, we follow the original concept of UED, which aims to learn a minimax regret policy \cite{minimax_regret} by generating training environments that maximize the regret of the agent. Based on this concept, the learning-based methods train an environment generator via reinforcement learning. PAIRED \cite{paired} estimates the regret with a difference between returns obtained by two distinct agents, and trains RL-based generator by utilizing the regret as a reward. Recently, CLUTR \cite{clutr} and SHED \cite{shed} utilize generative models to improve the performance of PAIRED. CLUTR trains the environment generator on the learned latent space, and SHED supplies the environment generator with augmented experiences created by diffusion models. Despite the progress, training the generator via RL is unstable due to the moving manifold \cite{clutr, gradientgamefail} and often struggles to generate diverse environments. On the other hand, replay-based methods based on PLR \cite{plr} utilize a random environment generator and decide which environments to replay. ACCEL \cite{accel} combines the evolutionary approaches \cite{poet, poet2} and PLR by taking random mutation on replayed environments. While these replay-based methods show scalable performance on a large-scale domain \cite{human_openended} and outperform the learning-based methods, they do not have the ability to directly generate meaningful environments. Unlike prior UED methods, we augment the regret objective of UED with an entropy regularization term and propose a method that employs a diffusion model as an environment generator to enhance the environment generation capability. Our work is also closely related to data augmentation for training robust policy. Particularly, DRAGEN \cite{dragen} and ISAGrasp \cite{shape_augmentation} augment existing data in learned latent spaces to train a policy that is robust to unseen scenarios. Our algorithm, on the other hand, focuses on generating curricula of environments without any prior knowledge and dataset.

\subsection{Diffusion Models}
Diffusion models \cite{ddpm, score, ddim} have achieved remarkable performance in various domains, such as image generation \cite{imagen}, video generation \cite{videodiffusion}, and  robotics \cite{diffuser, dtamp, policydiffusion}. Particularly, diffusion models effectively perform conditional generation using guidance to generate samples conditioned on class labels \cite{classifier_guidance, classifier_free} or text inputs \cite{latentdiffusion, dalle, video2}. Prior works also guide the diffusion models utilizing an additional network or loss functions, such as adversarial guidance to generate images to attack a classifier \cite{advdiffuser}, safety guidance using pre-defined functions to generate safety-critical driving scenarios \cite{drivingscene}, and guidance using reward functions trained by human preferences to generate censored samples. \cite{censored_sampling}. We note that our implementation of the regret-guided diffusion model is based on the work of Dhariwal et al. \cite{classifier_guidance} and Yoon et al. \cite{censored_sampling}.

\section{Background}
\subsection{Unsupervised Environment Design}
\label{background:ued}
Unsupervised environment design (UED, \cite{paired}) aims to provide an adaptive curriculum to learn a policy that successfully generalizes to various environments. 
The environments are represented using a Markov decision process (MDP), defined as $\langle A, S, \mathcal{P, R}, \rho_0, \gamma \rangle$, where A is a set of actions, S is a set of states, $\mathcal{P}: S \times A \times S \rightarrow \left[0, 1\right]$ is a transition model, $\mathcal{R}: S \times A \rightarrow \mathbb{R}$ is a reward function, $\rho_0: S \rightarrow \left[0, 1\right]$ is an initial state distribution and $\gamma$ is a discount factor. UED employs an environment generator that designs environments by controlling free environment parameters of underspecified environments, which is represented using an underspecified Markov decision process (UMDP). UMDP is defined as $\mathcal{M} = \langle A, S, \Theta, \mathcal{P^M, R^M}, \rho_0^{\mathcal{M}}, \gamma \rangle$, where $\Theta$ is a set of free environment parameters. Assigning a value to the free environment parameter $\theta \in \Theta$ results in a specific MDP $\mathcal{M_\theta}$ with the environment configuration ($\mathcal{P^\theta = P^M(\theta), R^\theta = R^M(\theta)}, \rho_0^\theta = \rho_0^{\mathcal{M}}(\theta)$).
For example, when learning a mobile robot to navigate towards the goal point while avoiding obstacles, $\theta$ could represent the positions of obstacles, the position of the goal, and the start position of the robot.

UED algorithms alternate between designing a set of environments and training the agent on the generated environments. The environment generator first produces an environment parameter $\theta$ that maximizes the agent's regret. The regret of the policy $\pi$ on environment $\mathcal{M}^{\theta}$ is defined as,
\begin{equation}
\label{regret definition}
    \regret(\pi, \theta) := - V(\pi, \theta) + \max_{\pi' \in \Pi} V(\pi', \theta),
\end{equation}
where $\Pi$ is a set of policies and $V(\pi, \theta) := \mathbb{E}_{\rho_0^\theta, \pi, \mathcal{P}^\theta}\left[\sum_{n=0}^N r_n\gamma^n\right]$ is an expected return where $r_n$ is a reward obtained by $\pi$ at timestep $n$ on $\mathcal{M}^\theta$. Then, the agent is trained on the generated environment to maximize the expected return, resulting in minimizing the regret. This framework can be formulated with the following two-player minimax game:
\begin{equation}
\label{standard ued}
    \min_{\pi \in \Pi}\,\max_{\theta \in \Theta}{\regret(\pi, \theta)}.
\end{equation}
It is ensured that the agent learns the minimax regret policy $\pi^* \in \mathop{\mathrm{argmin}}\limits_{\pi \in \Pi}\,\max\limits_{\theta \in \Theta}\regret(\pi, \theta)$ by assuming the two-player game (\ref{standard ued}) reaches the Nash equilibrium \cite{paired, repaired}. However, learning the minimax regret policy is challenging. Since the objective (\ref{standard ued}) does not guarantee the diversity of generated environments, the agent may not be trained on sufficiently various environments. 

\subsection{Diffusion Probabilistic Models}
A diffusion probabilistic model \cite{ddpm} is a generative model that generates samples from noise via iterative denoising steps. Diffusion models start with perturbing data by progressively adding Gaussian noise, called the \textit{forward process}. The forward process can be modeled with a value-preserving stochastic differential equation (VP SDE, \cite{score}):
\begin{equation}
\label{forward sde}
    dX_{t} = -\frac{\beta_{t}}{2}X_{t}dt + \sqrt{\beta_{t}}dW_{t},
\end{equation}
where $t \in \left[0, T\right]$ is a continuous diffusion time variable, $\beta_t > 0$ is a variance schedule, and $W_{t}$ is a standard Brownian motion. Since the forward process (\ref{forward sde}) has tractable conditional marginal distributions $p_{t}(X_{t} \vert X_0) = \mathcal{N}(\sqrt{\alpha_{t}}X_0, (1 - \alpha_{t})I)$ where $\alpha_{t} = e^{- \int_{0}^{t} \beta_{t} \, dt}$, $p_T(X_T)$ will be corrupted into $\mathcal{N}(0,I)$ when $T \rightarrow \infty$. 

Generating samples following the data distribution $p_{data}(\cdot)$ requires a \textit{reverse process}, a reverse-time SDE that has the same marginal distributions as the forward process (\ref{forward sde}). By Anderson's theorem \cite{anderson1982reverse}, the reverse process can be formulated with a reverse-time SDE defined as,
\begin{equation}
\label{backward sde}
    d{X}_{t} = -\beta_{t}\left[\frac{1}{2}{X}_{t} + \nabla_{\scalebox{.6}{$X_t$}}\log p_{t}({X}_{t})\right]dt + \sqrt{\beta_{t}}d{W}_{t}.
\end{equation}
Hence, learning a diffusion model means learning a score network $s_{\phi}(X_{t}, t)$ that approximates a score function $\nabla_{\scalebox{.6}{$X_t$}}\log p_{t}({X}_{t})$. The score network is trained via score matching \cite{scorematching}, then plugged into the reverse process (\ref{backward sde}):
\begin{equation}
\label{backward approx sde}
    d{X}_{t} = -\beta_{t}\left[\frac{1}{2}{X}_{t} + s_{\phi}(X_{t}, t)\right]dt + \sqrt{\beta_{t}}d{W}_{t}.
\end{equation}
Indeed, we can generate samples by solving the approximated reverse process (\ref{backward approx sde}) with an initial condition $X_{T} \sim \mathcal{N}(0, I)$. 

To generate samples with label $Y$ using the diffusion model, the score function of the conditional distribution $p_{t}(X_{t} \vert Y)$ should be estimated. Since $p_{t}(X_{t} \vert Y) \propto p_{t}(X_{t}, Y) = p_{t}(X_{t})p_{t}(Y \vert X_{t})$ due to Bayes' rule, conditional samples can be generated by solving the reverse process with classifier guidance \cite{classifier_guidance}:
\begin{equation}
\begin{aligned}
\label{classifier guidance}
    \hat{s}_{\phi}(X_{t}, t) & =s_{\phi}(X_{t}, t) + \omega\nabla_{\scalebox{.6}{$X_t$}}\log \hat{p}_{t}(Y \vert X_{t}),\\
    d{X}_{t} & =-\beta_{t}\left[\frac{1}{2}{X}_{t} + \hat{s}_{\phi}(X_{t}, t)\right]dt + \sqrt{\beta_{t}}d{W}_{t},
\end{aligned}
\end{equation}
where $\hat{p}_{t}(Y \vert X_{t})$ is a time-dependent classifier network and $\omega > 0$ is a guidance weight to scale classifier gradients.

\section{Proposed Method}
In this section, we describe our approach to employ a diffusion model as an environment generator to enhance the environment generation capability. We first introduce soft UED, which mutates UED to be more suitable for using a diffusion model as a generator by augmenting the original objective with the entropy regularization term. Then, we propose a novel soft UED algorithm, adversarial environment design via regret-guided diffusion models (ADD). ADD consists of two key components: 1) a diffusion-based environment generation by using the regret as a guidance, and 2) a method to estimate the regret in a differentiable form. We present these key components in detail and conclude the section with an explanation of the overall system and its advantages compared to prior UED methods. 

\subsection{Soft Unsupervised Environment Design}
In this section, we introduce soft UED, designed to guarantee the diversity of environments by adding an entropy regularization term to the original UED objective (\ref{standard ued}). Soft UED is defined as the following minimax game between the agent and the environment generator: 
\begin{equation}
\label{soft ued}
    \min_{\pi \in \Pi}\,\max_{\Lambda \in \mathcal{D}_\Lambda} \mathop{\mathbb{E}}_{\theta \sim \Lambda}\left[\regret(\pi, \theta)\right] + \frac{1}{\omega}H(\Lambda),
\end{equation}
where $\Lambda$ is a distribution over $\Theta$, $\mathcal{D}_\Lambda$ is a set of distributions over $\Theta$, $H(\Lambda) := -\sum\limits_{\theta \in \Theta}\Lambda(\theta)\log\Lambda(\theta)$ is an entropy of $\Lambda$, and $\omega$ is a regularization coefficient. Based on the fact that $H$ is concave, we can show that the strong duality holds:
\begin{proposition}
\label{soft_ued_saddlepoint}
Let $L(\pi, \Lambda):=\mathop{\mathbb{E}}_{\theta \sim \Lambda}\left[\regret(\pi, \theta)\right] + \frac{1}{\omega}H(\Lambda)$ and assume that $S, A,$ and $\Theta$ are finite. Then, 
$\min\limits_{\pi \in \Pi}\,\max\limits_{\Lambda \in \mathcal{D}_\Lambda} L(\pi, \Lambda) = \max\limits_{\Lambda \in \mathcal{D}_\Lambda}\,\min\limits_{\pi \in \Pi} L(\pi, \Lambda)$. 
\end{proposition}
The proof is detailed in Appendix \ref{appendix:saddlepoint}. Proposition \ref{soft_ued_saddlepoint} implies that there exists a valid optimal point $(\tilde{\pi}, \tilde{\Lambda})$, and it is stationary for alternative optimization of $\pi$ and $\Lambda$. Hence, the agent will learn a soft minimax regret policy $\tilde{\pi} = \mathop{\mathrm{argmin}}\limits_{\pi \in \Pi}\max\limits_{\Lambda \in \mathcal{D}_\Lambda} L(\pi, \Lambda)$ if it reaches the optimal point.
One of the most significant difference from the original UED is the role of the environment generator. Instead of finding a specific environment parameter that maximizes the regret, soft UED updates the environment generator to sample environment parameters from a distribution that maximizes the objective function of soft UED.

We note that the soft UED framework encompasses prior UED algorithms. In the learning-based methods, the generator is trained with RL using an entropy bonus, which is known to enhance performance \cite{entropy_paired} and plays a similar role to $H(\Lambda)$. The replay-based methods also consider the diversity of environments by sampling environment parameters from a probability distribution proportional to the regret, instead of selecting a parameter that maximizes the regret. Therefore, soft UED can be considered as a general framework that incorporates practical methods.

\subsection{Regret-Guided Diffusion Models}
Soft UED converts the problem of generating regret-maximizing environments into a problem of sampling the environment parameter $\theta$ from a desired distribution $\Lambda^{\pi} := \mathop{\mathrm{argmax}}\limits_{\Lambda \in \mathcal{D}_\Lambda} L(\pi, \Lambda)$. It is a well-known fact that $\Lambda^\pi$ has a closed-form solution as follows:
\begin{equation}
\label{add objective}
    \Lambda^{\pi}(\theta) = \frac{u(\theta) \exp(\omega \regret(\pi, \theta))}{C^{\pi}},
\end{equation}
where $C^\pi$ is a normalizing constant, and $u(\cdot)$ denotes an uniform distribution over $\Theta$. Inspired by the classifier guidance (\ref{classifier guidance}), we solve this sampling problem by guiding a pre-trained diffusion model with the regret. To this end, we decompose the score function of $\Lambda^\pi$ as follows:
\begin{equation}
\label{regret guidance}
    \nabla_{\scalebox{.6}{$\theta_t$}} \log \Lambda^{\pi}_t(\theta_t) = \nabla_{\scalebox{.6}{$\theta_t$}} \log u_t(\theta_t) + \omega \nabla_{\scalebox{.6}{$\theta_t$}} \regret_t(\pi, \theta_t),
\end{equation}
where $t$ is a diffusion time variable, $\theta_t$ is an environment parameter perturbed by the forward process (\ref{forward sde}), $u_t(\cdot)$ denotes a distribution of $\theta_t$ when $\theta_0 \sim u(\cdot)$, $\Lambda^{\pi}_t(\cdot)$ denotes a distribution of $\theta_t$ when $\theta_0 \sim \Lambda^{\pi}(\cdot)$, and $\regret_t(\pi, \theta_t)$ is a time-dependent regret on the noised environment $\theta_t$, which is equal to $\regret(\pi, \theta_0)$. We approximate the first term $\nabla_{\scalebox{.6}{$\theta_t$}} \log u_t(\theta_t)$ with a score network $s_\phi(\theta_t, t) \approx \nabla_{\scalebox{.6}{$\theta_t$}} \log u_t(\theta_t)$ that is learned by training a diffusion-based environment generator on the randomly generated environment dataset before the agent begins to learn. Then, we can formulate a regret-guided reverse process with a reverse-time SDE as follows:
\begin{equation}
\begin{aligned}
\label{regret guided reverse process}
    s^{\pi}_{\phi}(\theta_{t}, t) & =s_{\phi}(\theta_{t}, t) + \omega\nabla_{\scalebox{.6}{$\theta_t$}} \regret_t(\pi, \theta_t),\\
    d{\theta}_{t} & =-\beta_{t}\left[\frac{1}{2}{\theta}_{t} + s^{\pi}_{\phi}(\theta_{t}, t)\right]dt + \sqrt{\beta_{t}}d{W}_{t}.
\end{aligned}
\end{equation}

Hence, if a gradient of the regret is tractable, we can sample an environment parameter $\theta_0$ from the desired distribution $\Lambda^{\pi}$ by solving the regret-guided reverse process (\ref{regret guided reverse process}) with an initial condition $\theta_T \sim \mathcal{N}(0, I)$. 
However, the regret (\ref{regret definition}) is intractable since we cannot access the environment-specific optimal policy. Prior works on UED propose various methods to estimate the regret using episodic returns or temporal difference errors, but none of them are differentiable w.r.t. $\theta_t$ since agents cannot access the environment parameter and the reward function. 

\subsection{A Differentiable Regret Estimation}
In order to estimate the regret in a differentiable form, we present a novel method based on a flexible regret \cite{paired}, which is known to enhance the performance of the learning-based UEDs \cite{clutr}. The main idea is to estimate the regret with a difference between the maximum and average episodic returns that can be achieved by the agent. To make it differentiable w.r.t. $\theta_t$, we utilize an environment critic that predicts the return of the agent in the given environment parameter, as done in DSAGE \cite{deep_surrogate} and LPG \cite{jackson2023discovering}. The environment critic $\tau_\psi$ learns to predict a distribution of returns, analogous to distributional RL \cite{c51}, to better capture the stochasticity of the environment and policy. Based on a support defined as $\{z_i = v_{min} + \frac{i}{M-1}(v_{max} - v_{min})\}_{i=0}^{M-1}$, which is a set of centers of M bins that evenly divide the return domain $\left[v_{min}, v_{max}\right]$, we obtain an estimated categorical return distribution from an environment critic output $l(\theta_t, t;\psi) \in \mathbb{R}^M$ as follows:
\begin{equation}
\label{return distribution}
    \hat{\mathcal{Z}_\pi}(\theta_t, t) = z_i \quad \text{w.p.} \quad \frac{\text{exp}(l_i(\theta_t, t;\psi))}{\sum_{j=0}^{M-1}\text{exp}(l_j(\theta_t, t;\psi))}.
\end{equation}
To align $\hat{\mathcal{Z}}_\pi$ with a true return distribution, we train the environment critic by gradient descent on the cross entropy loss between $\hat{\mathcal{Z}_\pi}(\theta_t, t)$ and a target distribution, which is constructed by projecting episodic returns that the agent achieves on the environment $\mathcal{M}^{\theta_0}$ onto the support $\{z_i\}_{i=0}^{M-1}$. 

After the environment critic is updated, we estimate the regret (\ref{regret definition}) with a difference between the maximum return that the current agent can achieve and average of the predicted return distribution. However, the process of finding a maximum achievable return from the distribution is not differentiable. To address this issue, we further approximate the maximum with a conditional value at risk (CVaR), based on the fact that $\text{CVaR}_\alpha(\hat{\mathcal{Z}})$ converges to the maximum as a risk measure $\alpha$ goes zero. As a result, we estimate the regret of the agent as follows:
\begin{equation}
\label{differentialbe regret}
    \regret_t(\theta_t, t) \approx \text{CVaR}_{\alpha}(\hat{\mathcal{Z}_\pi}(\theta_t, t)) - \mathbb{E}(\hat{\mathcal{Z}_\pi}(\theta_t, t)).
\end{equation}

\begin{figure}[t]
\includegraphics[width=0.75\textwidth]{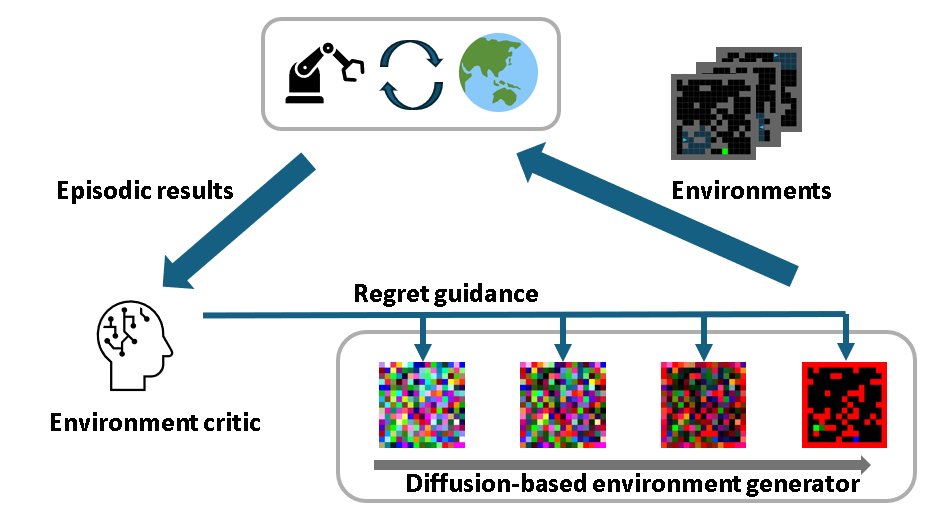}
\centering
\caption{\textbf{Overview of ADD.} After the agent is trained on environments produced by the environment generator, the environment critic is updated using the episodic results. Then, the environment critic guides the diffusion-based environment generator with the regret to produce adversarial environments. By repeating this process, the agent learns a policy that is robust to environmental changes.}
\label{overview}
\end{figure}

\subsection{Adversarial Environment Design via Regret-Guided Diffusion Models}
\label{method:overview}
An overview of ADD is provided in Figure \ref{overview}. First, a diffusion-based environment generator, which is pre-trained on the randomly generated environment dataset, produces a set of environments for the agent. After the agent interacts with the generated environments and is trained via reinforcement learning, the episodic results are utilized to update the environment critic. Then, the environment critic estimates the regret of the agent in a differentiable form (\ref{differentialbe regret}) and guides the reverse process of the diffusion-based environment generator (\ref{regret guided reverse process}), resulting in environment parameters following the distribution that maximizes the soft UED objective (\ref{soft ued}). By repeating this process, the agent learns the soft minimax regret policy, which is robust to the variations of environments. A pseudocode of the algorithm is shown in Appendix \ref{appendix:psuedocode}.

Since ADD does not require an additional training of the pre-trained diffusion model, the ability to cover the high-dimensional environment domain can be preserved. Furthermore, ADD enables the generator to directly produce meaningful environments via regret-guided reverse process. Therefore, ADD can be seen as having both the advantage of the replay-based methods and learning-based methods while resolving some of their limitations. Additionally, we can control the difficulty level of the generated environments after the training of the RL agent is over. In detail, we can generate environments of difficulty level $k \in \{1,2,\dots ,M\}$ by replacing the regret in the regret-guided reverse process (\ref{regret guided reverse process}) with a log probability of achieving a specific return $z_{M-k}$ as follows:
\begin{equation}
\begin{aligned}
\label{controldiffreverseprocesss}
s_{\phi}'(\theta_{t}, t) &=s_{\phi}(\theta_{t}, t) + \omega\nabla_{\scalebox{.6}{$\theta_t$}} \log \text{Pr}(\hat{\mathcal{Z}_\pi}(\theta_t, t)=z_{M-k}),\\
    d{\theta}_{t} & =-\beta_{t}\left[\frac{1}{2}{\theta}_{t} + s_{\phi}'(\theta_{t}, t)\right]dt + \sqrt{\beta_{t}}d{W}_{t}.
\end{aligned}
\end{equation}
It enables the reuse of the learned generator and environment critic in various applications, such as generating benchmarks with varying difficulties. 

\section{Experiments}
\label{Experiments}

\textbf{Tasks} We conduct extensive experiments with two challenging tasks. First, we evaluate the proposed method on a partially observable navigation task with a discrete action space and sparse rewards. Then, we assess the performance of our algorithm on a 2D bipedal locomotion task, which has a continuous action space while offering dense rewards.

\textbf{Baselines} We compare the proposed method against several UED baselines. For the learning-based method, we use PAIRED \cite{paired}, which trains the environment generator via reinforcement learning. For the replay-based method, we use PLR$^{\perp}$ \cite{repaired}, which utilizes the random generator and updates the agent only with episodes from the replayed environments. To benchmark performance, we use ACCEL \cite{accel}, a current state-of-the-art UED algorithm that applies random mutations to replayed environments. Among the two implementation methods of the ACCEL, we use the one that samples environment parameters from the full parameter range, rather than the one that restricts the sampling to a range that ensures simple environments are generated, as the latter could be seen as incorporating prior knowledge. Domain randomization (DR) is also included in baselines so that we can demonstrate the effectiveness of UED. Lastly, we use ADD w/o guidance to show whether the regret guidance induces the diffusion model to generate adversarial environments and enhances the performance of the agent. 

\textbf{Outline} We first train a diffusion-based environment generator on the randomly generated environment dataset. Then, we use proximal policy optimization (PPO, \cite{ppo}) to train the agent on the environments generated by UED methods. To evaluate the generalization capability of the trained agent, we measure the zero-shot transfer performance in challenging, human-designed environments. Additionally, to understand where the differences in performance originate, we conduct quantitative and qualitative analyses on the curriculum of the generated environments by tracking complexity metrics and drawing t-SNE plots. For space consideration, we elaborate on detailed experimental settings including environment parameterization methods in Appendix \ref{appendix:experiments}. 

\subsection{Partially Observable Navigation Task}
We first evaluate the proposed method on a maze navigation task \cite{paired}, which is based on the Minigrid \cite{minigrid}. In this task, an agent is trained to take a discrete action using an observation from its surroundings to receive a sparse reward upon reaching a goal. For prior UED methods, we set the maximum number of blocks in a grid environment to 60, aligning with Parker-Holder et al. \cite{accel}. For the proposed method, we train the diffusion-based environment generator on 10M random environments whose number of blocks uniformly varies from zero to 60. Then, we train the LSTM-based policy for 250M environmental steps and evaluate the zero-shot performance on 12 challenging test environments from prior works \cite{paired, repaired}. 

\begin{figure}[t]
\includegraphics[width=1.0\textwidth]{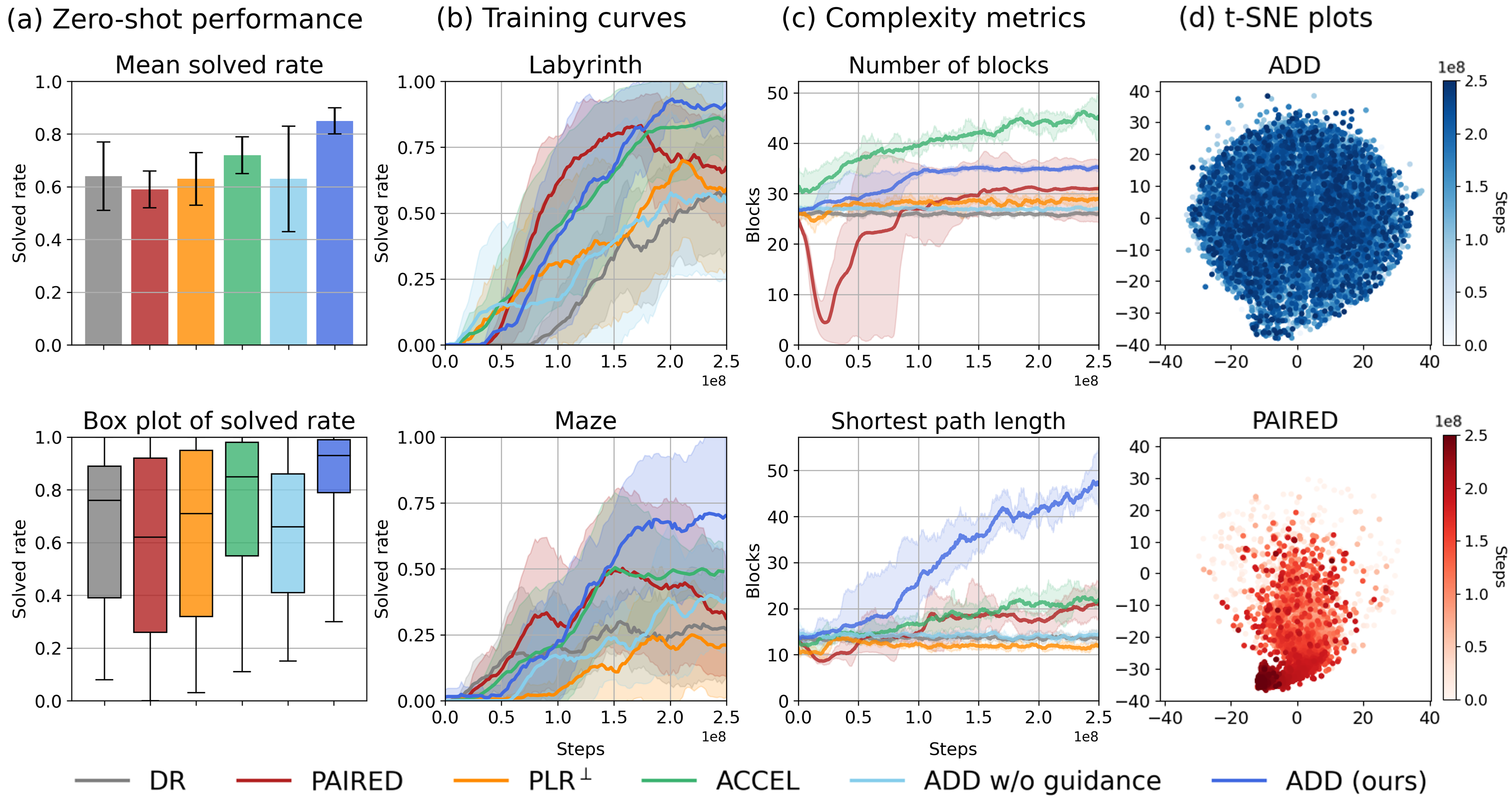}
\centering
\caption{\textbf{Partially observable navigation results.} \textbf{(a)}: Zero-shot performance on the 12 test environments. We report results across five random seeds, each evaluated over 100 independent episodes per environment. \textbf{(b)}: Training curves on two challenging test environments. \textbf{(c)}: Complexity metrics of the generated environments during training. \textbf{(d)}: t-SNE embedding of the generated environments during training.}
\label{maze_result}
\vspace{-5pt}
\end{figure}

\textbf{Performance.} In Figure \ref{maze_result}(a), we report the mean solved rate and box plot to compare the zero-shot performance of the learned policy on the test environments. The result demonstrates that ADD outperforms the baseline methods while achieving 85\% of the mean solved rate, which is 18\% higher compared to the ACCEL. Furthermore, ADD achieves the highest Q1, Q2, and Q3 values while its interquartile range, defined as Q3 - Q1, is 51\% smaller than ACCEL. Therefore, we can infer that the proposed method consistently outperforms the baselines in the challenging test environments. In Figure \ref{maze_result}(b), we report training curves on two test environments consisting of Maze and Labyrinth. The results demonstrate that ADD shows the monotonic performance improvement and achieves a higher solved rate compared to other baselines. While ACCEL shows 13\% higher mean solved rate than DR, PAIRED and PLR$^{\perp}$ do not show notable performance improvement by applying UED techniques. Particularly, PAIRED shows 8\% lower mean solved rate compared to DR, demonstrating the challenge of the learning-based UED methods. ADD w/o guidance shows 63\% of mean solved rate similar to DR, demonstrating that the regret guidance is critical for training the robust policy. Please refer to Appendix \ref{appendix:mazeresult} for detailed per-environment results. 

We note that the performance is measured after the fixed number of environmental steps, in line with some UED papers \cite{repaired, dred} and traditional deep reinforcement learning research. In contrast, Parker-Holder et al. \cite{accel} recorded performance after the fixed number of policy updates. Since the replay-based methods require additional episodes without policy updates, using the number of environmental steps may be seen as unfair to PLR$^{\perp}$ and ACCEL. To address this issue, we also measured the performance of our method trained with only half the environmental steps, aligning the number of policy updates with PLR$^{\perp}$ and ACCEL. When using half the environmental steps, ADD achieves a 72\% mean solved rate, which ties with ACCEL. This demonstrates that the proposed method remains competitive, even when using the number of policy updates as a metric.

\textbf{Generated curriculum.} In Figure \ref{maze_result}(c), we report complexity metrics consisting of the number of blocks and shortest path length. The results demonstrate that complexity metrics of ADD w/o guidance and DR are almost consistent over time since they do not consider the policy. While PLR$^{\perp}$ eventually generates environments with a larger number of blocks compared to DR, ADD and PAIRED generate a curriculum with significantly increasing complexity. Specifically, ADD eventually generates environments with the second largest number of blocks and the longest shortest path. ACCEL also shows significantly increasing complexities despite being based on PLR$^{\perp}$. This is because only up to 60 blocks exist on the 13X13 grid, so random mutation increases the expectation of the number of blocks. Therefore, it can be seen that ADD and PAIRED efficiently generate complex environments by adapting to the current policy while PLR$^{\perp}$ struggles to find environments with high complexity. To compare the distribution of generated environments, we report t-SNE \cite{tsne} plots in Figure \ref{maze_result}(d). While the environments generated by PAIRED eventually cover only a specific region, the environments generated by ADD cover a significantly larger region over time. The results demonstrate that ADD successfully generates adversarial environments while preserving diversity.

\subsection{2D Bipedal Locomotion Task}
We evaluate the proposed method on the 2D bipedal locomotion task, which is based on the BipedalWalker task in OpenAI Gym \cite{gym} and adopted by Parker Holder et al. \cite{accel}. In this task, an agent is trained to control its four motors using observation from its Lidar sensor and joints to walk over challenging terrain. UED methods, including the proposed algorithm, need to provide environment parameters consisting of stump height, stair height, pit gap, stair steps, and ground roughness. We note that each parameter increases the difficulty of the environment as its value increases. We train the RL agent for two billion environmental steps and evaluate the zero-shot performance on six test environments. 


\begin{figure}[t]
\includegraphics[width=1.0\textwidth]{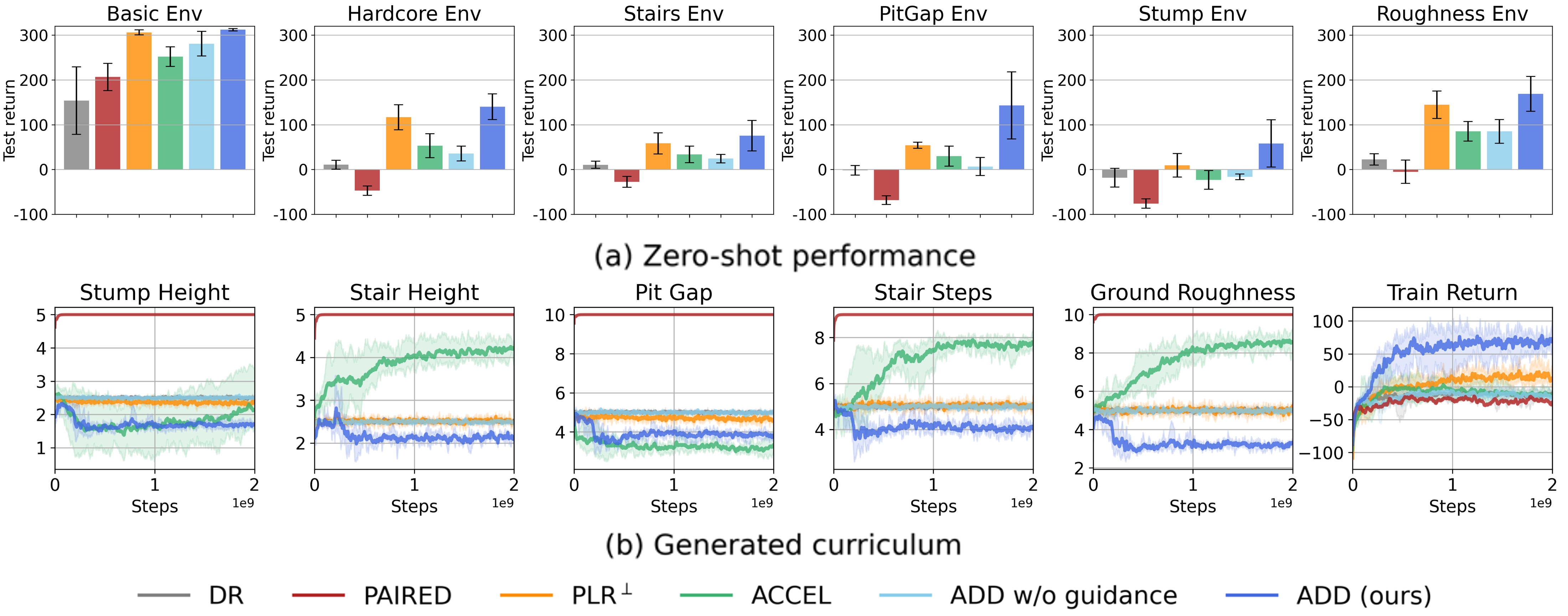}
\centering
\caption{\textbf{2D bipedal locomotion task results.} \textbf{(a)}: Zero-shot performance on the six test environments. We report results across five random seeds, each evaluated over 100 independent episodes per environment. \textbf{(b)}: Complexity metrics of the generated environments and episodic return achieved during training.}
\label{bipedal_result}
\vspace{-5pt}
\end{figure}

\textbf{Performance.} Figure \ref{bipedal_result}(a) shows the average return on each test environment. The proposed algorithm achieves the highest return across all environments, with an average of 149.6. Even with half the environmental steps, it achieves a score of 127.4, still surpassing PLR$^\perp$. ACCEL shows lower performance than PLR$^\perp$, which can be attributed to the lower sample efficiency induced by the additional interaction between the environment and the agent to assess the modified environments. On the other hand, PAIRED achieves the lowest return in all test environments except the easiest Basic Environment. This shows that the learning-based methods struggle to train a robust policy in practice. We note that a recent work \cite{entropy_paired} stabilizes the training of PAIRED in this task by integrating the evolutionary concept of ACCEL. While applying the evolutionary approach to ADD is possible, we leave it for future work. Lastly, ADD w/o guidance demonstrates superior generalization performance compared to DR. Although these two methods are theoretically identical, this difference is presumably caused by the limited size of the dataset used for training the diffusion-based environment generator.

\textbf{Generated curriculum} Figure \ref{bipedal_result}(b) presents the complexity metrics of the generated training environments and the episodic returns achieved by the RL agent. Unlike the partially observable navigation task, the complexity measure of the environments generated by ADD gradually decreases. This result arises since the randomly generated environments are excessively challenging for the current agent. As evidence, examining the returns achieved by the agent in the generated environments reveals that all methods, except for ADD, consistently yield returns of 0 or below. From these results, we can infer that the proposed algorithm generates environments that are not merely more difficult but are conducive to the agent's learning process. For detailed experimental results including a qualitative analysis on the generated environments, please refer to Appendix \ref{appendix:bipedalresult}.

\begin{figure}[t!]
\includegraphics[width=1.0\textwidth]{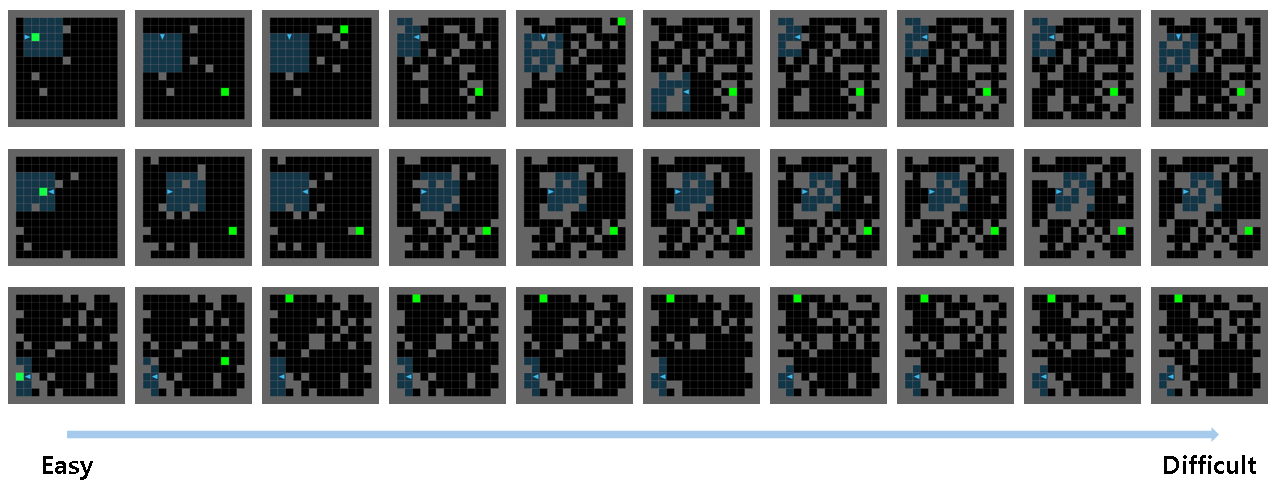}
\centering
\caption{\textbf{Controllable generation results for the partially observable navigation task}. The figure shows the results of guiding the generator to generate progressively more difficult environments. We note that each row is generated from the same initial noise $\theta_T$.}
\label{maze_controllable_generation}
\vspace{-5pt}
\end{figure}

\subsection{Controllable Generation}
To demonstrate the ability to control the difficulty level of the generated environments, we provide the example environment generation results in Figure \ref{maze_controllable_generation}. We control the difficulty level $k$ by guiding the diffusion-based environment generator with a log probability of achieving a specific return $z_{M-k}$, as described in (\ref{controldiffreverseprocesss}). We vary $k$ from zero to $M-1$ so that the difficulty level of generated environments increases. Environments generated with $k=0$, which are shown in the leftmost images, include fewer blocks and a close proximity between the agent's starting position and the goal. As k increases, environments are generated with a greater number of blocks and a larger distance between the starting position and the goal, resulting in the elimination of all possible paths when $k=M-1$. The results demonstrate that we can effectively control the difficulty level of the environment using the diffusion-based environment generator and learned environment critic, without domain knowledge. We also present the results of controlling difficulty levels for the 2D bipedal locomotion task in Appendix \ref{appendix:bipedalresult}.

\section{Limitation}
\label{limitation}
While the proposed method is suitable for training a robust policy, there exist several limitations. First, despite the existence of the optimal point is proven in Proposition \ref{soft_ued_saddlepoint}, convergence to such optimal point is not guaranteed.
Furthermore, the difference between the true value of the regret and its estimate is not tightly bounded. Lastly, updating the environment critic using episodic results cannot exactly capture the current agent's capability since the policy is updated after the episode. Hence, exploring methods to estimate the regret with a rigorous theoretical background would be an interesting topic for future work.

\section{Conclusion}
In this work, we present a novel UED algorithm that exploits the representation power of diffusion models. We first introduce soft UED, which augments the original UED objective with an entropy regularization term to make it suitable for using a diffusion-based environment generator. Then, we propose ADD, which guides the pre-trained diffusion model with a novel regret estimator to produce environments that are conducive to train a robust policy. Our experimental results demonstrate that ADD is capable of training a policy that successfully generalizes to challenging environments. Moreover, it has been verified that ADD generates an instructive curriculum with varying complexity while covering large environment configuration spaces.

\section*{Acknowledgments and Disclosure of Funding}
This work was partly supported by Institute of Information \& Communications Technology Planning \& Evaluation (IITP) grant funded by the Korea government (MSIT) (No. 2019-0-01190, [SW Star Lab] Robot Learning: Efficient, Safe, and Socially-Acceptable Machine Learning, 40\%), Basic Science Research Program through the National Research Foundation of Korea (NRF) funded by the Ministry of Science and ICT (NRF-2022R1A2C2008239, General-Purpose Deep Reinforcement Learning Using Metaverse for Real World Applications, 40\%), and Institute of Information \& communications Technology Planning \& Evaluation (IITP) grant funded by the Korea government(MSIT) (NO.RS-2021-II211343, Artificial Intelligence Graduate School Program [Seoul National University], 20\%).

\bibliography{main}
\bibliographystyle{unsrt}

\newpage

\appendix

\section{Algorithm Details}
\label{appendix:algo}

\subsection{Proof of Proposition \ref{soft_ued_saddlepoint}}
\label{appendix:saddlepoint}
In this section, we show that the minimax problem of soft UED (\ref{soft ued}) has zero minimax duality gap. We assume that $S, A,$ and $\Theta$ are finite to avoid the technical issues regarding compactness of a set of distributions. Following Section \ref{background:ued}, we denote a reward function, transition probability, and initial state distribution of an environment $\mathcal{M}^\theta$ with $\mathcal{R}^\theta, \mathcal{P}^\theta,$ and $ \rho_0^\theta$, respectively.

We first define an occupancy measure, for a policy $\pi \in \Pi$ and an environment $\mathcal{M}^\theta$, as $\rho_\pi^\theta(s,a) = \pi(a\vert s)\sum_{n=0}^{N}\gamma^n\,\text{Pr}(s_n=s \vert \pi, \theta)$. Then, there is a one-to-one correspondence between $\Pi$ and a set of valid occupancy measures $\mathcal{D}^\theta := \{\rho: \sum_a\rho(s,a) = \rho_0^\theta(s) + \gamma\sum_{s', a'} \mathcal{P}^\theta(s \vert s', a')\rho(s', a') \}$ \cite{apprenticeship, gail}. If we define a global occupancy measure as $\rho_\pi := \left[\rho_\pi^{\theta^i}\right]_{i=1}^{\left|\Theta\right|}$, where $\theta^i$ is an ith element of $\Theta$, it is obvious that there is a one-to-one correspondence between $\Pi$ and a set of valid global occupancy measures $\mathcal{D} \subset \mathcal{D}^{\theta^1} \times \cdots \mathcal{D}^{\theta^{\left|\Theta\right|}}$. Therefore, we can replace $\pi$ in the objective function of soft UED (\ref{soft ued}) with $\rho_\pi$, as in the following lemma:
\begin{lemma}
\label{occupancy_objective_validation}
if $\bar{L}(\rho_\pi, \Lambda) = \sum\limits_{\theta} (V^*(\theta) - \sum\limits_{s, a} \rho_{\pi}^ \theta(s, a)\mathcal{R}^{\theta}(s,a))\Lambda(\theta) + \frac{1}{\omega}H(\Lambda)$, where $V^*(\theta) = \max\limits_{\pi^{A} \in \Pi}V(\pi^A, \theta)$, then $\bar{L}(\rho_\pi, \Lambda) = L(\pi, \Lambda)$.
\end{lemma}
\begin{proof}
Based on the definition of the regret (\ref{regret definition}), we have
\begin{equation}
\begin{aligned}
\label{lemma a1 proof}
    L(\pi, \Lambda) & = \mathop{\mathbb{E}}_{\theta \sim \Lambda}\left[\regret(\pi, \theta)\right] + \frac{1}{\omega}H(\Lambda)\\
     & = \sum\limits_{\theta} \regret(\pi, \theta) \Lambda(\theta) + \frac{1}{\omega}H(\Lambda)\\
     & = \sum\limits_{\theta} (V^*(\theta) - V(\pi, \theta))\Lambda(\theta) + \frac{1}{\omega}H(\Lambda)\\
     & = \sum\limits_{\theta} (V^*(\theta) - \sum\limits_{s, a} \rho_{\pi}^ \theta(s, a)\mathcal{R}^{\theta}(s,a))\Lambda(\theta) + \frac{1}{\omega}H(\Lambda)\\
     & = \bar{L}(\rho_\pi, \Lambda).
\end{aligned}
\end{equation}
\end{proof}

Now, we can rewrite the objective function of soft UED with a global occupancy measure as follows:
\begin{equation}
\label{occupancy minimax game}
\min\limits_{\rho \in \mathcal{D}}\,\max\limits_{\Lambda \in \mathcal{D}_\Lambda}\bar{L}(\rho, \Lambda).
\end{equation}
Then, we can prove Proposition \ref{soft_ued_saddlepoint} by showing (\ref{occupancy minimax game}) has zero duality gap. However, we cannot apply minimax theorem \cite{sion_minimax} directly since $\mathcal{D}$ is not a convex set. To resolve this issue, we first augment the problem as follows:
\begin{equation}
\label{augmented problem}
\min\limits_{\rho \in \bar{\mathcal{D}}}\,\max\limits_{\Lambda \in \mathcal{D}_\Lambda}\bar{L}(\rho, \Lambda),
\end{equation}
where $\bar{\mathcal{D}} := \{\sum_{k=1}^K w_k\rho_k \vert  K \in \mathbb{N},\sum_{k=1}^Kw_k = 1, \forall k \in \{1, \dots, K\}: w_k \geqq 0, \rho_k \in \mathcal{D}\}$ is a convex hull of $\mathcal{D}$. We will show that the augmented problem (\ref{augmented problem}) has zero minimax duality gap, and end the proof by showing the optimal values of the augmented problem can also be reached by the original problem (\ref{occupancy minimax game}).

\begin{lemma}
\label{augmented_minmax_duality}
$\min\limits_{\rho \in \bar{\mathcal{D}}}\,\max\limits_{\Lambda \in \mathcal{D}_\Lambda}\bar{L}(\rho, \Lambda) = \max\limits_{\Lambda \in \mathcal{D}_\Lambda}\,\min\limits_{\rho \in \bar{\mathcal{D}}}\bar{L}(\rho, \Lambda)$
\end{lemma}
\begin{proof}
Since $\bar{L}(\rho, \Lambda)$ is a linear combination of $\rho^\theta$, it is convex for all $\rho$. Furthermore, $\bar{L}(\rho, \Lambda)$ is concave for $\Lambda$ since the entropy $H$ is concave. Therefore, based on the fact that $\bar{\mathcal{D}}$ and $\mathcal{D}_\Lambda$ are both convex and compact, the augmented problem has zero duality gap due to minimax theorem \cite{sion_minimax}.
\end{proof}

\begin{lemma}
\label{convex_hull_to_original}
For every $\Lambda \in \mathcal{D}_\Lambda$ and corresponding $\rho^*(\Lambda) \in \mathop{\mathrm{argmin}}\limits_{\rho \in \bar{\mathcal{D}}} \bar{L}(\rho, \Lambda)$, there exists $\rho' \in \mathcal{D}$ such that $\bar{L}(\rho^*(\Lambda), \Lambda) = \bar{L}(\rho', \Lambda)$.
\end{lemma}
\begin{proof}
For every $\Lambda \in \mathcal{D}_\Lambda$ and corresponding $\rho^*(\Lambda) \in \mathop{\mathrm{argmin}}\limits_{\rho \in \bar{\mathcal{D}}} \bar{L}(\rho, \Lambda)$, there exist
$K \in \mathbb{N}, w_{1:K} \geqq 0$, and $\rho_{1:K} \in \mathcal{D}$ such that $\sum_{k=1}^K w_k = 1$ and $\rho^*(\Lambda) = \sum_{k=1}^K w_k\rho_k$. Then, following inequality holds:
\begin{equation}
\min \{\bar{L}(\rho_k, \Lambda)\}_{k=1}^K \geqq \bar{L}(\rho^*(\Lambda), \Lambda) = \sum_{k=1}^K w_k\bar{L}(\rho_k, \Lambda) \geqq \min \{\bar{L}(\rho_k, \Lambda)\}_{k=1}^K,
\end{equation}
where first inequality holds due to the definition of $\rho^*(\Lambda)$, equality holds since $\bar{L}$ is linear for $\rho$, and second inequality holds since $\rho^*(\Lambda)$ is a convex combination of $\rho_{1:K}$. Then, $\rho' \in \mathop{\mathrm{argmin}}\limits_{\rho \in \{\rho_k\}_{k=1}^K}\bar{L}(\rho, \Lambda)$ is an element of $\mathcal{D}$ and satisfies $\bar{L}(\rho^*(\Lambda), \Lambda) = \bar{L}(\rho', \Lambda)$.
\end{proof}
Lemma \ref{convex_hull_to_original} ensures the minimum value of $\bar{L}$ achieved over $\bar{\mathcal{D}}$ is also achievable over $\mathcal{D}$, implying the optimal value is the same for both the original and augmented problems. It confirms that strong duality holds for the original problem (\ref{occupancy minimax game}) as well.
\begin{proposition}
\label{occupancy_duality}
$\min\limits_{\rho \in \mathcal{D}}\,\max\limits_{\Lambda \in \mathcal{D}_\Lambda}\bar{L}(\rho, \Lambda) = \max\limits_{\Lambda \in \mathcal{D}_\Lambda}\,\min\limits_{\rho \in \mathcal{D}}\bar{L}(\rho, \Lambda)$
\end{proposition}

Hence, using Lemma \ref{occupancy_objective_validation} and Proposition \ref{occupancy_duality}, we can prove Proposition \ref{soft_ued_saddlepoint}:
\begin{equation}
\label{occupancy duality to policy}
    \min_{\pi \in \Pi}\,\max_{\Lambda \in \mathcal{D}_\Lambda}L(\pi, \Lambda) = \min_{\rho \in \mathcal{D}}\,\max_{\Lambda \in \mathcal{D}_\Lambda}\bar{L}(\rho, \Lambda) = \max_{\Lambda \in \mathcal{D}_\Lambda}\,\min_{\rho \in \mathcal{D}}\bar{L}(\rho, \Lambda) = \max_{\Lambda \in \mathcal{D}_\Lambda}\,\min_{\pi \in \Pi}L(\pi, \Lambda)
\end{equation}

\subsection{Diffusion Models}
\label{appendix:diffusion}
In this section, we present implementation details on diffusion models. To solve the forward and reverse processes (\ref{forward sde}, \ref{backward sde}), we follow the implementation of DDPM \cite{ddpm}, which can be viewed as a discretization of VP SDE (\ref{forward sde}). As as result, the forward process (\ref{forward sde}) is implemented as follows:
\begin{equation}
\label{ddpm forward process}
\theta_{t} = \sqrt{1 - \beta_t}\theta_{t-1} + \sqrt{\beta_t}Z_t,
\end{equation}
where $Z_t \sim \mathcal{N}(0, I)$. In the appendix, we make a slight abuse of notation by considering $t$ as a discrete variable to provide detailed implementation specifics. To solve the reverse process (\ref{backward sde}), we utilize a error network $\epsilon_\phi(\theta_t, t) = - \sqrt{1 - \alpha_t}s_\phi(\theta_t, t)$ instead of using $s_\phi$, where $\alpha_t = \prod_{t'=1}^t 1 - \beta_{t'}$, and $\beta_t$ follows the linear noise schedule. The error network is trained to minimize the loss $J(\theta_0, t; \phi) := \mathbb{E}_{\epsilon \sim \mathcal{N}(0, I)} \lVert\epsilon - \epsilon_\phi(\sqrt{\alpha_t}\theta_0 + \sqrt{1 - \alpha_t}\epsilon)\rVert^2$. To futher accelerate the sampling, we apply DDIM sampling \cite{ddim} as follows:
\begin{equation}
\label{ddim reverse process}
\theta_{t-1} = \frac{1}{\sqrt{1 - \beta_t}}\theta_t - (\frac{\sqrt{1 - \alpha_t}}{\sqrt{1 - \beta_t}} - \sqrt{1 - \frac{\alpha_t}{1 - \beta_t}})\epsilon_\phi(\theta_t, t).
\end{equation}
Since DDIM sampling (\ref{ddim reverse process}) is deterministic, we can sample the environment parameter $\theta$ with $T'$ denoising steps, which is less than the original diffusion timestep $T$.

\subsection{Environment Critic Update}
\label{appendix:envcriticupdate}
In this section, we present details on updating the environment critic. After the interaction between the RL agent with a policy $\pi$ and an environment $\mathcal{M}_\theta$, we construct a target return distribution $\mathcal{Z}_\pi^{\text{target}}(\theta)$ using episodic returns. Specifically, if episodic returns that the RL agent achieves are $\{v_k\}_{k=0}^{K-1}$, we give equal probabilities to each return and project them into the support $\{z_i\}_{i=0}^{M-1}$. As a result, the target return distribution is constructed as follows:
\begin{equation}
\label{appendix:target return distribution}
    \mathcal{Z}_\pi^{\text{target}}(\theta) = z_i \quad \text{w.p.} \quad \frac{1}{K} \sum_{k=0}^{K-1} \left[1 - \frac{\vert v_k - z_i \vert}{\Delta}\right]_0^1,
\end{equation}
where $\left[\cdot\right]_0^1$ bounds its argument in the range $\left[0, 1\right]$, and $\Delta := \frac{(v_{max} - v_{min})}{M}$ is a width of each bin. Then, we train the environment critic to produce the return distribution close to the target distribution (\ref{appendix:target return distribution}). To that end, we first sample $t$ from $1, \dots, T$ and obtain $\theta_t$ by solving the forward process (\ref{forward sde}) with initial condition $\theta_0 = \theta$, then construct the estimated return distribution $\hat{\mathcal{Z}_\pi}(\theta_t, t)$ using an output of the environment critic $\tau_\psi(\theta_t, t)$. Then, $\psi$ is updated via gradient descent to minimize the cross entropy loss between $\hat{\mathcal{Z}_\pi}(\theta_t, t)$ and $\mathcal{Z}_\pi^{\text{target}}(\theta)$. To prevent overfitting, we store episodic results in a buffer, then sample the environment parameters and their corresponding target return distribution from the buffer for training the environment critic.

\subsection{Pseudocode of ADD}
\label{appendix:psuedocode}

\begin{algorithm}[h]
\caption{Adversarial Environment Design via Regret-Guided Diffusion Models}
\label{psuedocode}
\begin{algorithmic}
\STATE {\bfseries Input:} Policy network $\pi_\xi$, diffusion model $s_\phi$, and environment critic network $\tau_\psi$.
\STATE Initialize network parameters $\xi,\phi, \psi$.
\STATE Train the diffusion model $s_\phi$ on a dataset of randomly generated environments.
\FOR{each epochs}
    \STATE Sample a set of environment parameters $\{\theta^i\}_{i=1}^B$ via regret-guided reverse process (\ref{regret guided reverse process}), whose regret is estimated using the environment critic (\ref{differentialbe regret}).
    \STATE Run episode on a set of environments $\{\mathcal{M}^{\theta^i}\}_{i=1}^B$ and update the policy $\pi_\xi$ via RL.
    \STATE Update the environment critic $\tau_\psi$ using episodic returns (Appendix \ref{appendix:envcriticupdate}).
\ENDFOR
\end{algorithmic}
\end{algorithm}

\section{Experiment Details and Hyperparameters}
\label{appendix:experiments}
In this section, we provide a comprehensive explanation of the experiments discussed in Section \ref{Experiments}. We begin by detailing the two tasks: partially observable navigation and 2D bipedal locomotion. Then, we conclude the section by reporting the hyperparameters employed in the experiments.
\subsection{Partially Observable Navigation Task}
\label{appendix:mazesetting}

\textbf{Environment details}. In the partially observable navigation task, which is based on the Minigrid \cite{minigrid} and adopted for UED in prior works \cite{paired}, the agent is trained to find and reach a goal in the grid maze environment. Each maze environment is a $15 \times 15$ grid whose cells on the edge are all walls, and the cells inside can contain walls, agents, or goals. When the agent reaches the goal, it receives a reward of $1 - N / N_{max}$, where $N$ is a length of the episode and $N_{max}$ is a maximum length of each episode. If it does not reach the goal, it receives a reward of 0. The agent uses a 7 × 7 grid around itself and its direction as an observation and chooses one of the actions: turn left, turn right, or go forward.

\begin{figure}[t]
\includegraphics[width=1.0\textwidth]{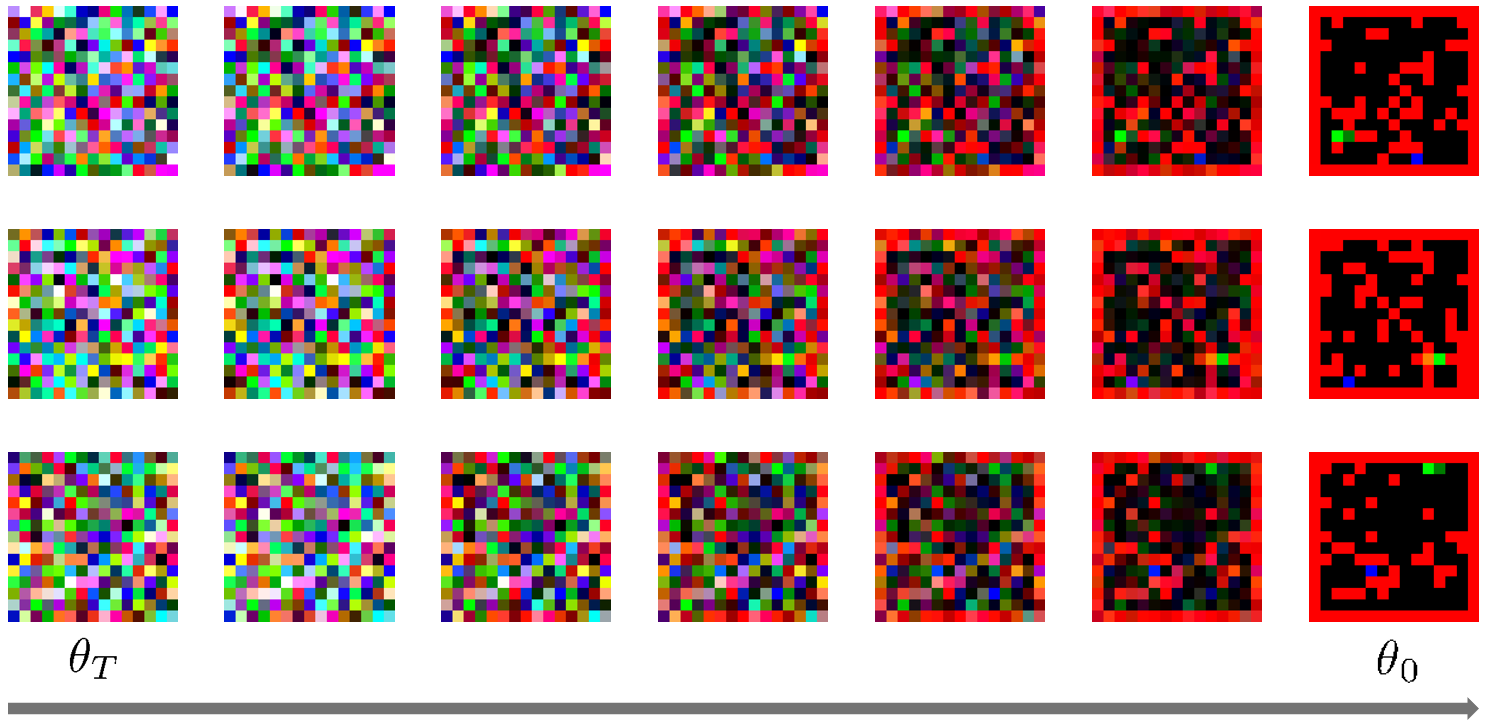}
\centering
\caption{\textbf{Maze environment generation using diffusion models.} We represent the maze environment with a parameter $\theta \in \mathbb{R}^{13 \times 13 \times 3}$, with each channel indicating the location of walls, the agent, and the goal. After training the diffusion-based environment generator on a dataset of randomly generated environment parameters, we can sample maze environments by solving the reverse process (\ref{backward approx sde}).}
\label{maze_diffusion}
\end{figure}

\textbf{Environment generation}. Aligning with Parker-Holder et al. \cite{accel}, we limit the number of walls that can exist in the 13 x 13 grid, excluding the walls on the edges, to 60. The RL generator of PAIRED selects locations to place walls over 60 steps, ensuring no changes if a wall already exists. After placing all the walls, it selects the starting position of the agent and the goal location. If a wall exists at those locations, it removes the wall and places the agent or the goal. On the other hand, the random generator used by PLR$^{\perp}$ and ACCEL uniformly samples the number of walls between 0 and 60 in advance and then choose the position to place the walls, agent and goal location randomly. This random generation is also used to create a dataset for training the diffusion-based environment generator of the proposed algorithm. Specifically, each environment parameter data is represented with a $13 \times 13 \times 3$ image. The first channel represents the location of the walls, with a value of one if a wall is present in the cell, and zero if it is not. The second channel indicates the starting position of the agent, with a value of one for the starting cell, 0.5 for the cell after moving forward once, and zero for all other cells. Finally, the third channel has a value of one for the cell corresponding to the goal, and zero otherwise. After training the diffusion-based environment generator on the randomly generated dataset, we can produce the environment parameter by employing the trained generator to solve reverse process (\ref{backward approx sde}), as shown in Figure \ref{maze_diffusion}.

\textbf{Training time}. All methods are trained utilizing RTX 3090Ti. To train DR and ADD w/o guidance, they require almost 48 hours to run 250 million environment steps. ADD requires almost 56 hours, and PLR$^\perp$ and ACCEL requires almost 100 hours for each random seed.

\subsection{2D Bipedal Locomotion Task}
\label{appendix:bipedalsetting}
\textbf{Environment details}. For the 2D bipedal locomotion task, we conduct the experiment using a modified version of the BipedalWalker environment from OpenAI Gym \cite{gym}, as done in Parker-Holder et al. \cite{accel}. In this task, the agent is trained to walk over challenging terrains by controlling four joints, and action is decided using a 24-dimensional observation, which is consisting of Lidar measurements, linear and angular velocities of the robot, positions and speeds of joints, and contact information. The agent receives a positive reward for moving forward, and receives -100 as a reward if it falls to the ground. If the agent reaches the opposite end of the terrain, total reward it receives is over 300. 

\textbf{Environment generation}. In the 2D bipedal locomotion task, we generate the environment by deciding the eight-dimensional environment parameter $\theta \in \mathbb{R}^8$, which is consisting of a min / max stump height, min / max stair height, min / max pit gap , stair steps, and roughness of the terrain. The RL generator of PAIRED selects each parameter sequentially, and the random generator of PLR$^{\perp}$ and ACCEL randomly decide each parameter by sampling a real number from its domain, which is reported in Table \ref{bipedal parameter}. We employ the random generator to construct a dataset of environment parameters, and train the diffusion-based environment generator, which will be used to produce the environment parameter while training the agent using the proposed method. After the environment parameter is decided, the entire environment is generated by the procedural content generation algorithm.

\textbf{Training time}. All methods are trained utilizing RTX 3090Ti. To train DR and ADD w/o guidance, they require almost 80 hours to run 2 billion environment steps. ADD requires almost 92 hours, and PLR$^\perp$ and ACCEL requires almost 160 hours to run the experiment for each random seed.

\renewcommand{\arraystretch}{1.2}
\begin{table}[hbt!]
\centering
\caption{\textbf{Domain of the environment parameter for the 2D bipedal locomotion task.}}
\label{bipedal parameter}
{%
\begin{tabular}{cccccc}
\hline
                & \textbf{Stump Height} & \textbf{Stair Height} & \textbf{Pit Gap} & \textbf{Stair Steps} & \textbf{Roughness} \\ \hline
\textbf{Domain} & $\left[0.0, 5.0\right]$                      & $\left[0.0, 5.0\right]$                      & $\left[0.0, 10.0\right]$                     & $\left[0.0, 10.0\right]$                   & $\left[1.0, 9.0\right]$                 \\[0.5ex] \hline
\end{tabular}%
}
\end{table}
\renewcommand{\arraystretch}{1}

\subsection{Hyperparameters}
\label{appendix:hyperparameters}
To train RL agents using UED baselines, we follow the implementation and hyperparameters of Parker-Holder et al. \cite{accel}, which is available at \url{https://github.com/facebookresearch/dcd}. The same parameters and network architecture were used to train the ADD agent, and the hyperparameters used are reported in the Table \ref{ppo parameter}. To train the diffusion-based environment generator, we followed the implementation of Dhariwal et al. \cite{classifier_guidance} and Yoon et al. \cite{censored_sampling}, which is available at \url{https://github.com/tetrzim/diffusion-human-feedback}, and the size of the randomly generated dataset is set to 10 million. For the partially observable navigation task, the architecture of the error network $\epsilon_\phi$ follows the UNet \cite{unet}, which utilizes three residual blocks with channel multipliers $\left[1, 2, 2, 2\right]$ for each resolution. For the 2D bipedal locomotion task, the architecture of error network is four-layer MLP with a sinusoidal time embedding. We report detailed hyperparameters used to train diffusion models in Table \ref{diffusion parameter}. Lastly, we report detailed hyperparameters for regret-guided diffusion process and training the environment critic in Table \ref{add parameter}. The network architecture of the environment critic is based on the UNet encoder for the partially observable navigation task and a four-layer MLP for the 2D bipedal locomotion task.
\begin{table}[hbt!]
\centering
\caption{\textbf{Hyperparameters used for training the RL agent in each task}}
\label{ppo parameter}
{%
\begin{tabular}{p{0.33\textwidth}|cc}
\hline
\textbf{Parameter} &\textbf{Minigrid}  &\textbf{BipedalWalker}  \\ \hline
$\gamma$ &0.995  &0.99  \\
$\lambda_\text{GAE}$ &0.95  &0.9  \\
PPO rollout length &256  &2000  \\
PPO epochs &5  &5  \\
PPO minibatches for epoch &1  &32  \\
PPO clip range &0.2  &0.2  \\
PPO number of workers &32  &16  \\
Adam learning rate &1e-4  &2e-4  \\
Adam $\epsilon$ &1e-5  &1e-5  \\
PPO max gradient norm &0.5  &0.5  \\
PPO value clipping &True  &False  \\
Return normalization &False  &True  \\
Value loss coefficient &0.5  &0.5  \\
Entropy coefficient &0.0  &1e-3  \\ 
LSTM-based policy &True  &False  \\ \hline
\end{tabular}%
}
\end{table}

\begin{table}[hbt!]
\centering
\caption{\textbf{Hyperparameters used for training the diffusion-based environment generator}}
\label{diffusion parameter}
{%
\begin{tabular}{p{0.33\textwidth}|cc}
\hline
\textbf{Parameter} &\textbf{Minigrid}  &\textbf{BipedalWalker}  \\ \hline
DDPM timestep $T$ &1000  &1000  \\
Network architecture &UNet  &MLP  \\
hidden dimension &128  &256  \\
Batch size &128  &512  \\
Dropout &0.0  &0.0  \\
AdamW learning rate &1e-4  &1e-4  \\ 
AdamW weight decay &0.05  &0.0  \\
AdamW $\beta_1$ &0.9  &0.9  \\
AdamW $\beta_2$ &0.999  &0.999  \\
EMA rate &0.9999  &0.9999  \\
Number of training steps &3e5  &1.5e5  \\ \hline
\end{tabular}%
}
\end{table}

\begin{table}[hbt!]
\centering
\caption{\textbf{Hyperparameters used for the regret guidance and training the environment critic}}
\label{add parameter}
{%
\begin{tabular}{p{0.33\textwidth}|cc}
\hline
\textbf{Parameter} &\textbf{Minigrid}  &\textbf{BipedalWalker}  \\ \hline
DDIM timestep $T'$  &50  &200  \\
Number of bins $M$ &100  &100  \\
Return domain $\left[v_{min}, v_{max}\right]$  &$\left[0, 1\right]$  &$\left[0, 300\right]$  \\
Guidance weight $\omega$ &5.0  &15.0  \\
CVaR risk level $\alpha$ &0.15  &0.3  \\
Environment critic minibatches &128  &128  \\
Environment ciritic epochs &5  &5  \\ 
Environment critic buffer size &1600 &800 \\ \hline
\end{tabular}%
}
\end{table}
\newpage
\section{Detailed Experimental Results}
\label{appendix:results}
In this section, we present detailed experimental results in the partially observable navigation task and 2D bipedal locomotion task. For each task, we will provide specific zero-shot generalization performance, t-SNE plots of all baselines, and the training environments generated by the proposed algorithm. Additionally, we will show examples of environments generated with varying difficulty levels using the method described in Section \ref{method:overview}.

\subsection{Partially Observable Navigation Task}
\label{appendix:mazeresult}
\begin{figure}[hbt!]
\includegraphics[width=1.0\textwidth]{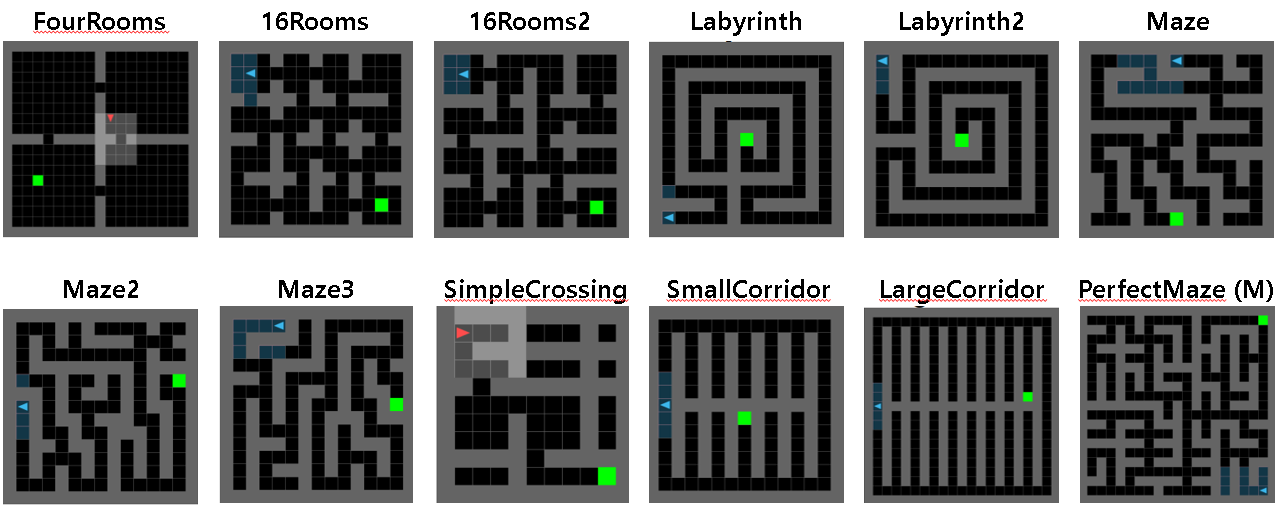}
\centering
\caption{\textbf{Zero-shot test environments for the partially observable navigation task.} SimpleCrossing and FourRooms environments are adopted from Chevalier-Boisvert et al. \cite{minigrid}, and other test environments are adopted from Dennis et al. \cite{paired} and Jiang et al. \cite{repaired, accel}.}
\label{maze_test_env}
\vspace{-2pt}
\end{figure}

\textbf{Zero-shot transfer test results.} After training the RL agent in the partially observable navigation task, we evaluate the generalization capability of the learned policy by testing the agent in twelve unseen environments, which are shown in Figure \ref{maze_test_env}. In each environment, the agent is evaluated for 100 independent episodes, and the full result is reported in Table \ref{maze_table}. We note that the reported quartile values represent the average of the quartile values of the solved rates achieved in the test environments for each seed. The results demonstrate that the proposed method achieves the best performance in five out of 12 test environments, with particularly high mean and quartile values compared to the baselines. Therefore, we can infer that ADD successfully trains an agent that is robust to environmental changes and generalizes to various environments.

\begin{table}[hbt!]
\caption{\textbf{Partially observable navigation task results.} The table shows the average solved rate and standard deviation over five independent runs, and each run is evaluated by 100 independent episodes for each test environment. All methods train the agent using LSTM-based PPO and evaluated after 250M environmental steps.}
\label{maze_table}
\resizebox{\textwidth}{!}{%
\begin{tabular}{p{2.5cm}|cccccc}
\hline
\hline
Environment     & DR                         & PAIRED             & PLR$^{\perp}$              & ACCEL                       & ADD w/o guidance           & ADD \\ \hline
FourRooms       & 0.62 $\pm\,$ 0.01          & 0.51 $\pm\,$ 0.05  & \textbf{0.64 $\pm\,$ 0.10} & 0.51 $\pm\,$ 0.05           & 0.61 $\pm\,$ 0.07          & 0.61 $\pm\,$ 0.09   \\
16Rooms         & 0.78 $\pm\,$ 0.21          & 0.96 $\pm\,$ 0.41  & 0.72 $\pm\,$ 0.09          & \textbf{0.97 $\pm\,$ 0.05}  & 0.93 $\pm\,$ 0.15          & 0.91 $\pm\,$ 0.13   \\
16Rooms2        & 0.50 $\pm\,$ 0.36          & 0.47 $\pm\,$ 0.28  & 0.60 $\pm\,$ 0.50          & 0.59 $\pm\,$ 0.41           & 0.63 $\pm\,$ 0.39          & \textbf{1.00 $\pm\,$ 0.10}   \\
Labyrinth       & 0.74 $\pm\,$ 0.42          & 0.74 $\pm\,$ 0.47  & 0.63 $\pm\,$ 0.42          & 0.96 $\pm\,$ 0.08           & 0.56 $\pm\,$ 0.41          & \textbf{1.00 $\pm\,$ 0.00}   \\
Labyrinth2      & 0.62 $\pm\,$ 0.49          & 0.49 $\pm\,$ 0.46  & 0.64 $\pm\,$ 0.50          & 0.73 $\pm\,$ 0.36           & 0.51 $\pm\,$ 0.54          & \textbf{0.97 $\pm\,$ 0.04}   \\
Maze            & 0.36 $\pm\,$ 0.47          & 0.06 $\pm\,$ 0.39  & 0.21 $\pm\,$ 0.08          & \textbf{0.82 $\pm\,$ 0.32}  & 0.18 $\pm\,$ 0.31          & 0.79 $\pm\,$ 0.37   \\
Maze2           & 0.46 $\pm\,$ 0.48          & 0.60 $\pm\,$ 0.17  & 0.18 $\pm\,$ 0.54          & \textbf{0.97 $\pm\,$ 0.03}  & 0.62 $\pm\,$ 0.44          & 0.76 $\pm\,$ 0.42   \\
Maze3           & \textbf{0.96} $\pm\,$ 0.09 & 0.64 $\pm\,$ 0.18  & 0.90 $\pm\,$ 0.42          & 0.61 $\pm\,$ 0.48           & 0.77 $\pm\,$ 0.38          & 0.74 $\pm\,$ 0.31   \\
SimpleCrossing  & 0.87 $\pm\,$ 0.06          & 0.82 $\pm\,$ 0.04  & \textbf{0.88 $\pm\,$ 0.09} & 0.75 $\pm\,$ 0.10           & 0.87 $\pm\,$ 0.02          & 0.80 $\pm\,$ 0.13   \\
SmallCorridor   & 0.63 $\pm\,$ 0.31          & 0.79 $\pm\,$ 0.02  & \textbf{0.97 $\pm\,$ 0.18} & 0.55 $\pm\,$ 0.46           & 0.84 $\pm\,$ 0.23          & 0.94 $\pm\,$ 0.09   \\
LargeCorridor   & 0.74 $\pm\,$ 0.35          & 0.43 $\pm\,$ 0.23  & 0.79 $\pm\,$ 0.36          & 0.56 $\pm\,$ 0.51           & 0.65 $\pm\,$ 0.33          & \textbf{0.95 $\pm\,$ 0.05}   \\
PerfectMaze (M) & 0.45 $\pm\,$ 0.20          & 0.43 $\pm\,$ 0.13  & 0.37 $\pm\,$ 0.24          & 0.64 $\pm\,$ 0.19           & 0.45 $\pm\,$ 0.21          & \textbf{0.82 $\pm\,$ 0.18}   \\ \hline
\textbf{Mean}   & 0.64 $\pm\,$ 0.13          & 0.59 $\pm\,$ 0.10  & 0.63 $\pm\,$ 0.07          & 0.72 $\pm\,$ 0.07           & 0.63 $\pm\,$ 0.20          & \textbf{0.85 $\pm\,$ 0.05}    \\ \hline
\textbf{First quartile}     & 0.39 $\pm\,$ 0.25          & 0.26 $\pm\,$ 0.25  & 0.32 $\pm\,$ 0.17          & 0.55 $\pm\,$ 0.15           & 0.41 $\pm\,$ 0.28          & \textbf{0.79 $\pm\,$ 0.07} \\
\textbf{Second quartile}     & 0.76 $\pm\,$ 0.19          & 0.62 $\pm\,$ 0.16  & 0.71 $\pm\,$ 0.13          & 0.85 $\pm\,$ 0.05           & 0.66 $\pm\,$ 0.30          & \textbf{0.93 $\pm\,$ 0.05} \\
\textbf{Third quartile}     & 0.89 $\pm\,$ 0.09          & 0.92 $\pm\,$ 0.08  & 0.95 $\pm\,$ 0.01          & 0.98 $\pm\,$ 0.01           & 0.86 $\pm\,$ 0.17          & \textbf{0.99 $\pm\,$ 0.01} \\ \hline\hline
\end{tabular}%
}
\end{table}

\begin{figure}[t!]
\includegraphics[width=1.0\textwidth]{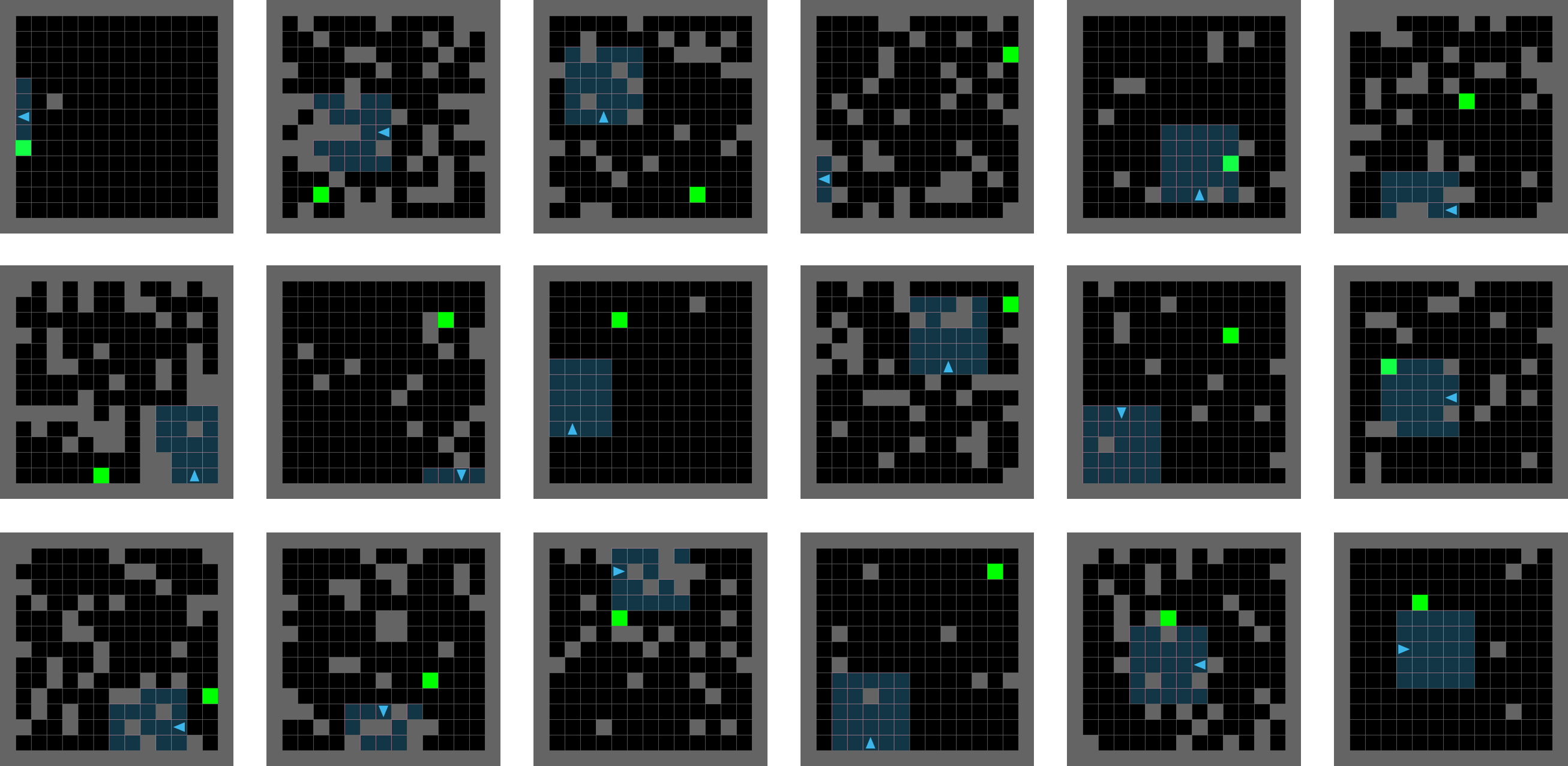}
\centering
\caption{\textbf{Examples of environments generated at the beginning of the agent's training in the partially observable task}. The figure shows 18 example environments that are generated right after the initiation of the agent's learning.}
\label{maze_first_20}
\end{figure}

\begin{figure}[t!]
\includegraphics[width=1.0\textwidth]{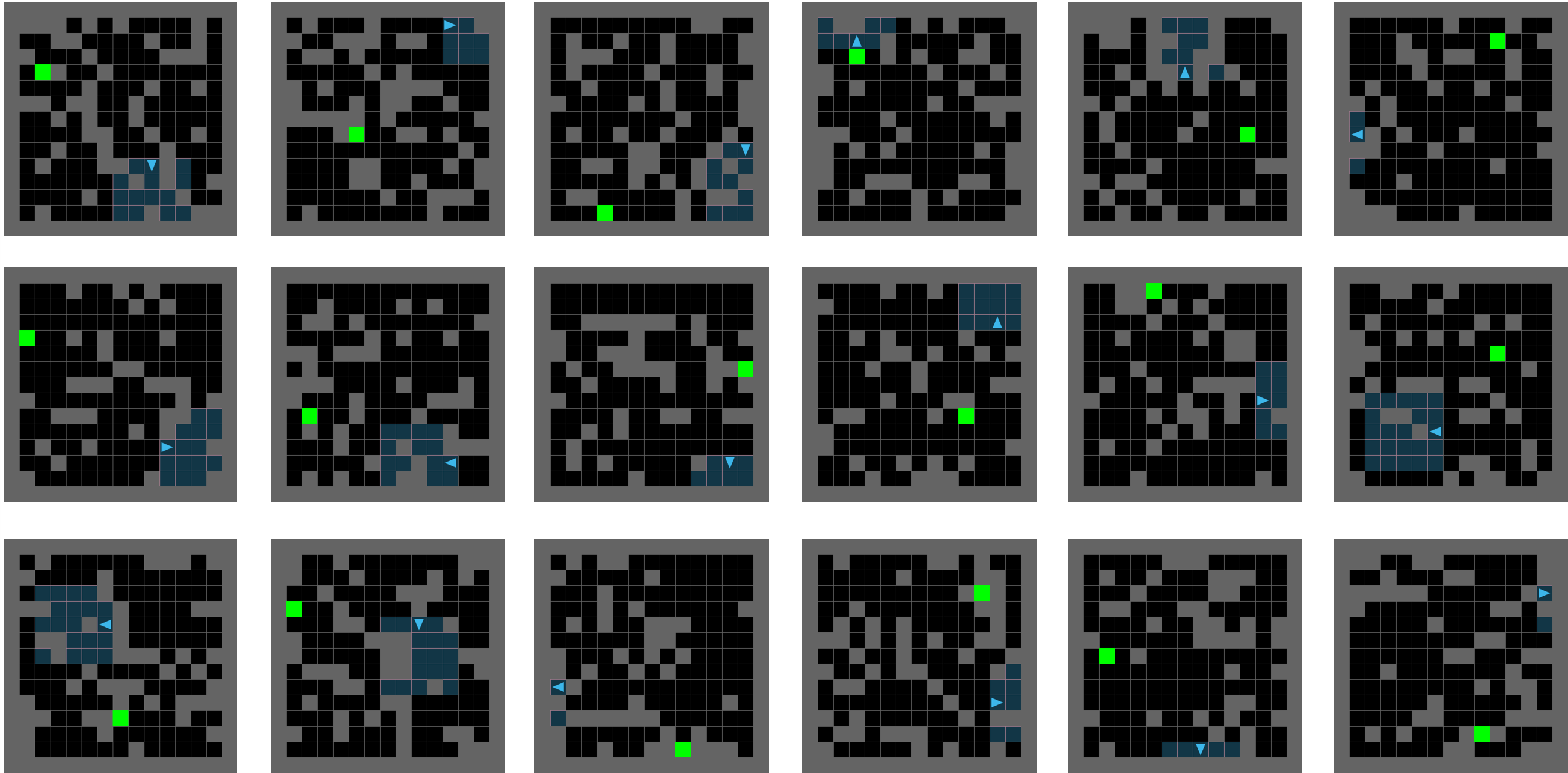}
\centering
\caption{\textbf{Examples of environments generated after 200 million environmental steps in the partially observable task}. The figure shows 18 example environments that are generated after 200 million environmental steps}
\label{maze_last_20}
\end{figure}

\newpage
\textbf{Generated training environments.} To support the claim regarding the generated curriculum, we provide example scenes of generated training environments in Figure \ref{maze_first_20} and Figure \ref{maze_last_20}. Comparing two figures reveals that the environments generated after training the agent with 200 million environmental steps are more complex and contain a larger number of blocks, aligning with the quantitative results shown in Figure \ref{maze_result}(c). The reason for this difference is that as the agent is trained, it achieves near-optimal performance in simple environments. Consequently, the environment critic predicts that the agent's regret will be larger in environments with a greater number of blocks and increased complexity.

\begin{figure}[t!]
\includegraphics[width=1.0\textwidth]{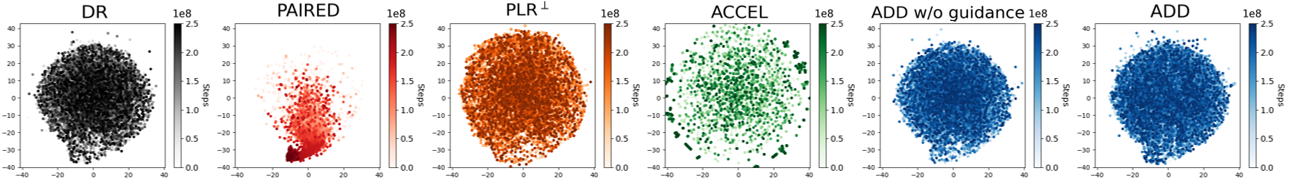}
\centering
\caption{\textbf{t-SNE plots of training environments in the partially observable navigation task}. The figure shows t-SNE plots of ADD and baselines. The results are obtained by mapping the environment parameters to a latent space using the encoder of the learned environment critic, followed by training a t-SNE.}
\label{maze_tsne_entire}
\end{figure}

\newpage
\textbf{t-SNE plots}. To conduct a qualitative analysis on the generated environments, we visualize training environments generated from ADD and baselines. Since the parameter of the maze environment is high-dimensional and hard to define the meaningful distance, we first utilize the encoder of the environment critic to map environment parameters to the learned latent space, then train a t-SNE on the latent vectors. The t-SNE results are shown in Figure \ref{maze_tsne_entire}. The results demonstrate that the proposed method generate sufficiently diverse environments comparable to those produced by a random generator.

\subsection{2D Bipedal Locomotion Task}
\label{appendix:bipedalresult}
\begin{figure}[hbt!]
\includegraphics[width=1.0\textwidth]{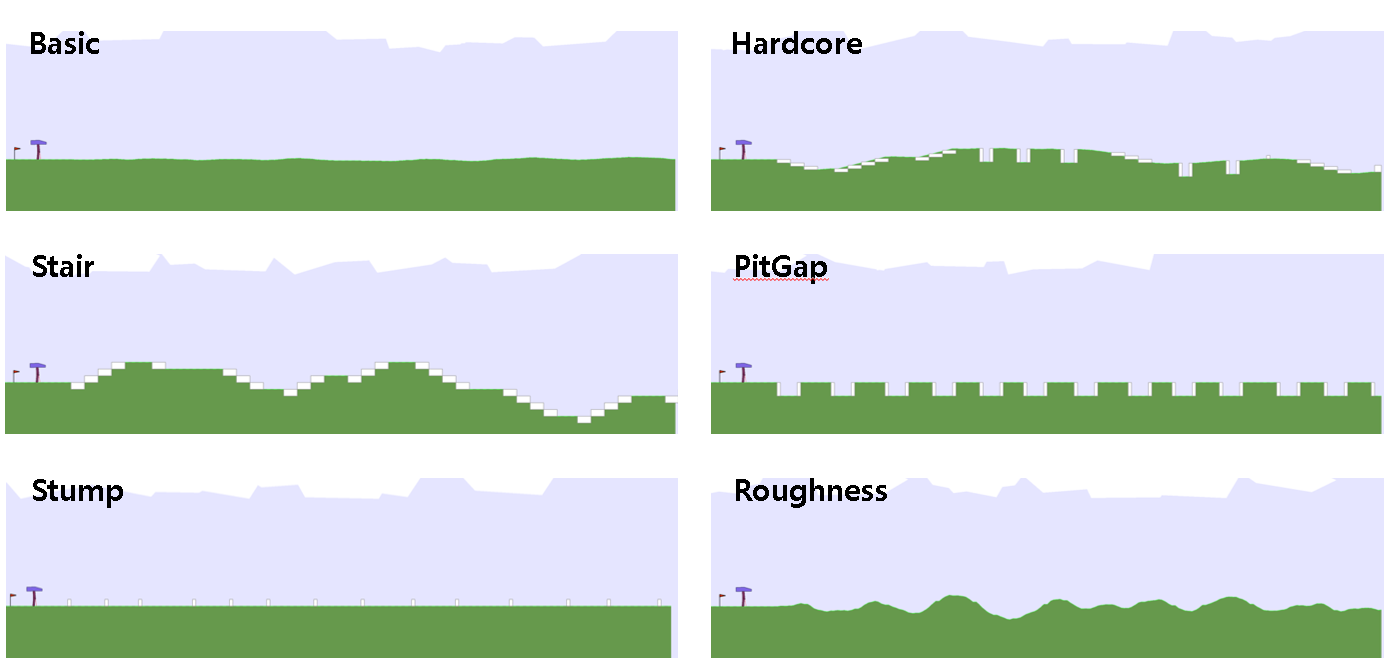}
\centering
\caption{\textbf{Test environments for the 2D bipedal locomotion task}. Basic and Hardcore environments are adopted from OpenAI Gym \cite{gym}, and other four test environments are adopted from Parker Holder et al. \cite{accel}.}
\label{bipedal_test_env}
\end{figure}

\textbf{Zero-shot transfer test results.} For the 2D bipedal locomotion task, we assess the zero-shot performance of the trained RL agent in six unseen test environments shown in Figure \ref{bipedal_test_env}. In each test environment, we evaluate the performance for 100 independent episodes. In this task, PAIRED struggles to train the RL-based environment generator since the value function often diverges. Therefore, we adopted the zero-shot performance result of PAIRED reported in Parker Holder et al. \cite{accel}. We report the full results in Table \ref{bipedal_table}, and the results demonstrate that the proposed method outperforms baselines across all test environments.

\begin{table}[hbt!]
\caption{\textbf{2D bipedal locomotion task results.} The table shows the average return and standard deviation over five independent runs, and each run is evaluated by 100 independent trials on each test environment. All methods train the agent using PPO and evaluated after two billion environmental steps.}
\label{bipedal_table}
\resizebox{\textwidth}{!}{%
\begin{tabular}{p{2.5cm}|cccccc}
\hline
\hline
Environment     & DR                  & PAIRED              & PLR$^{\perp}$       & ACCEL               & ADD w/o guidance   & ADD \\ \hline
Basic           & 153.7 $\pm\,$ 75.5  & 206.5 $\pm\,$ 30.3  & 306.0 $\pm\,$ 5.6   & 252.0 $\pm\,$ 21.9  & 280.8 $\pm\, 27.3 $           & \textbf{312.0 $\pm\,$ 2.5}   \\
Hardcore        & 10.6  $\pm\,$ 10.1  & -47.2 $\pm\,$ 10.6  & 116.6 $\pm\,$ 27.8  & 53.1  $\pm\,$ 26.6  & 35.5 $\pm\, 16.5$           & \textbf{140.1 $\pm\,$ 28.8}   \\
Stairs          & 10.5  $\pm\,$ 8.1   & -27.4 $\pm\,$ 12.1  & 58.4  $\pm\,$ 23.6  & 33.6  $\pm\,$ 18.5  & 24.3 $\pm\, 9.4$           & \textbf{75.4 $\pm\,$ 34.1}   \\
PitGap          & -1.8  $\pm\,$ 10.7  & -68.2 $\pm\,$ 9.7   & 54.2  $\pm\,$ 6.8   & 29.6  $\pm\,$ 22.4  & 6.7 $\pm\, 20.4$           & \textbf{143.2 $\pm\,$ 74.9}   \\
Stump           & -18.5 $\pm\,$ 20.9  & -76.0 $\pm\,$ 10.3  & 9.2   $\pm\,$ 26.3  & -23.2 $\pm\,$ 20.8  & -16.3 $\pm\, 6.3$           & \textbf{58.2 $\pm\,$ 52.6}   \\
Roughness       & 22.5  $\pm\,$ 12.5  & -5.1  $\pm\,$ 25.9  & 144.5 $\pm\,$ 30.8  & 85.2  $\pm\,$ 21.8  & 85.1 $\pm\, 26.6$           & \textbf{168.9 $\pm\,$ 38.8}   \\ \hline
\textbf{Mean}   & 29.5  $\pm\,$ 16.5  & -2.9  $\pm\,$ 14.5  & 114.8 $\pm\,$ 17.3  & 71.7  $\pm\,$ 15.9  & 69.3 $\pm\, 12.0$           & \textbf{149.6 $\pm\,$ 33.0}    \\ \hline \hline
\end{tabular}%
}
\end{table}

\begin{figure}[t!]
\includegraphics[width=1.0\textwidth]{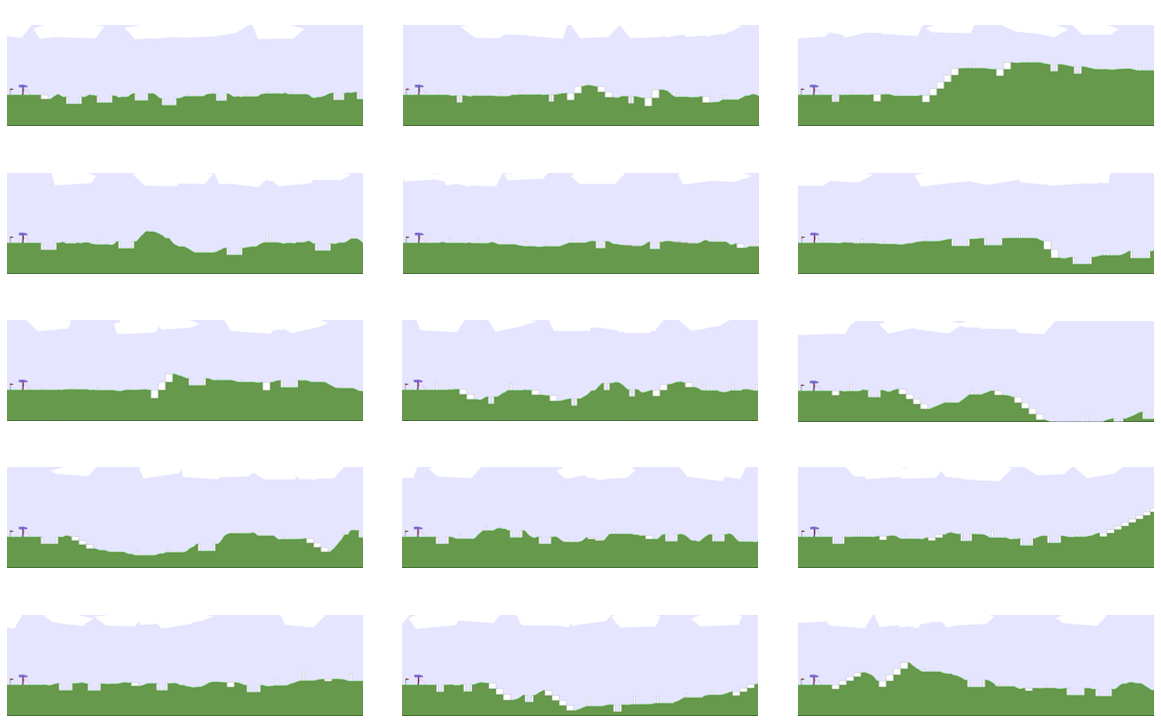}
\centering
\caption{\textbf{Examples of environments generated at the beginning of the agent's training in the 2D bipedal locomotion task}. The figure presents 15 example environments that are generated right after the agent starts learning.}
\label{bipedal_first_15}
\vspace{10pt}
\end{figure}

\begin{figure}[t!]
\includegraphics[width=1.0\textwidth]{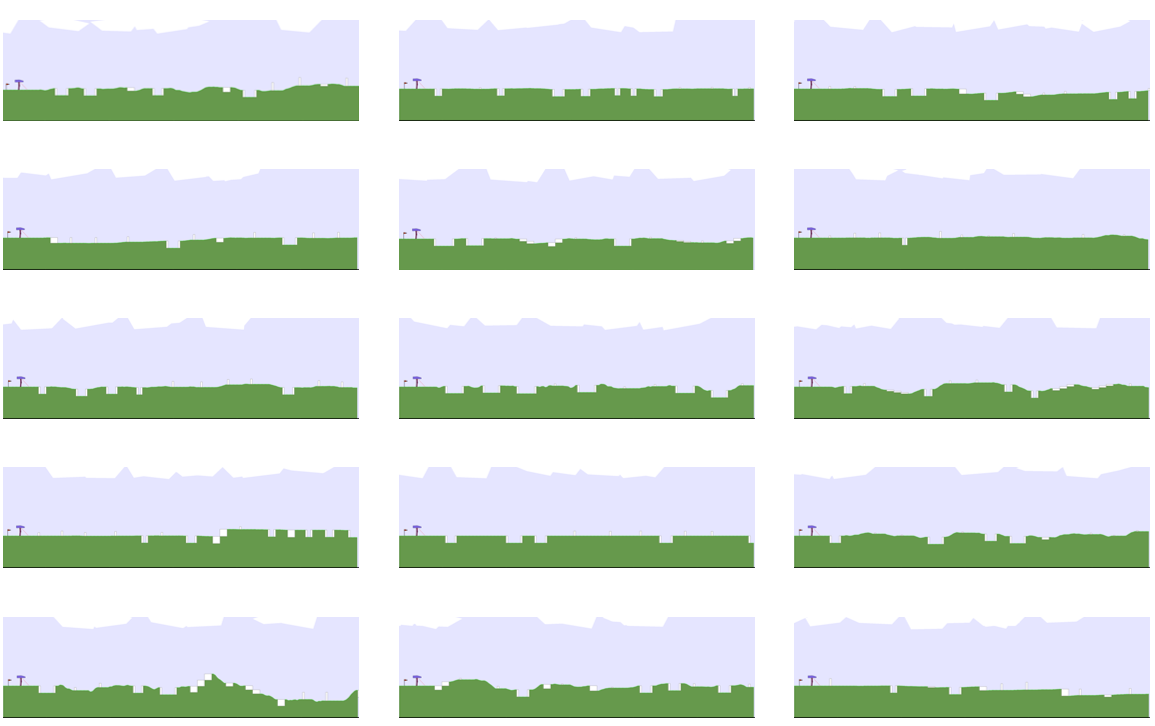}
\centering
\caption{\textbf{Examples of environments generated after 1.5 billion environmental steps in the 2D bipedal locomotion task}. The figure presents 15 example environments that are generated after 1,5 billion environmental steps.}
\label{bipedal_last_15}
\vspace{4pt}
\end{figure}

\newpage
\textbf{Generated training environments.} To demonstrate the generated environments vary as the agent learns, we provide the generated environments in the early and later stages of the training in Figure \ref{bipedal_first_15} and Figure \ref{bipedal_last_15}. It can be observed that the environments shown in Figure \ref{bipedal_last_15} are much simpler compared to those in Figure \ref{bipedal_first_15}, which is consistent with the quantitative analysis provided in Figure \ref{bipedal_result}(b). This difference occurs because randomly sampling environment parameters can result in overly challenging environments that hinder the agent's learning, and the environment critic guides the generator to produce simpler environments that are conducive to the agent's learning.

\clearpage

\begin{figure}[h]
\includegraphics[width=1.0\textwidth]{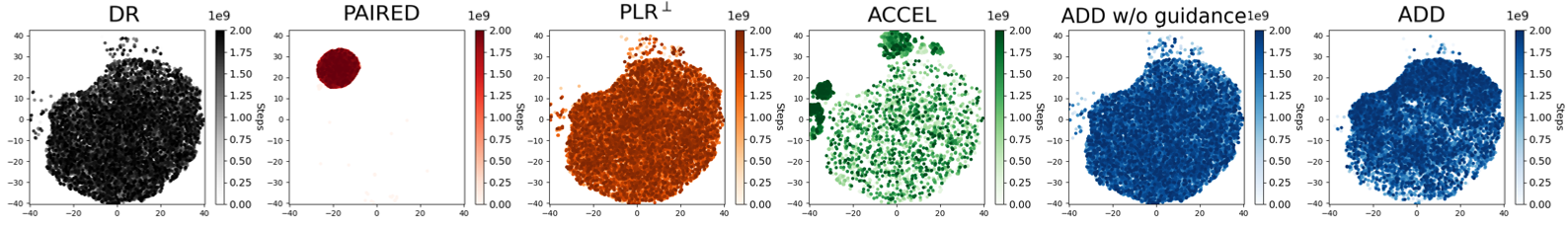}
\centering
\caption{\textbf{t-SNE plots of training environments in the 2D bipedal locomotion task}. The figure shows t-SNE plots of ADD and baselines, which are obtained by training a t-SNE to visualize environment parameters in the two-dimensional space.}
\label{bipedal_tsne_entire}
\end{figure}

\newpage
\textbf{t-SNE plots}. As in the previous task, we trained a t-SNE to visualize the generated environments for the 2D bipedal locomotion task. Since the environment parameters for this task have low dimensionality and the meaningful distance can be computed using L2 distance, we use the raw environment parameters without mapping them to a learned latent space. The results shown in Figure \ref{bipedal_tsne_entire} demonstrate that the proposed method produces sufficiently diverse training environments. In contrast, training environments generated by PAIRED are concentrated in a small area. Due to this difference, the agent trained with the proposed algorithm demonstrates better performance across all test environments.

\begin{figure}[h]
\includegraphics[width=1.0\textwidth]{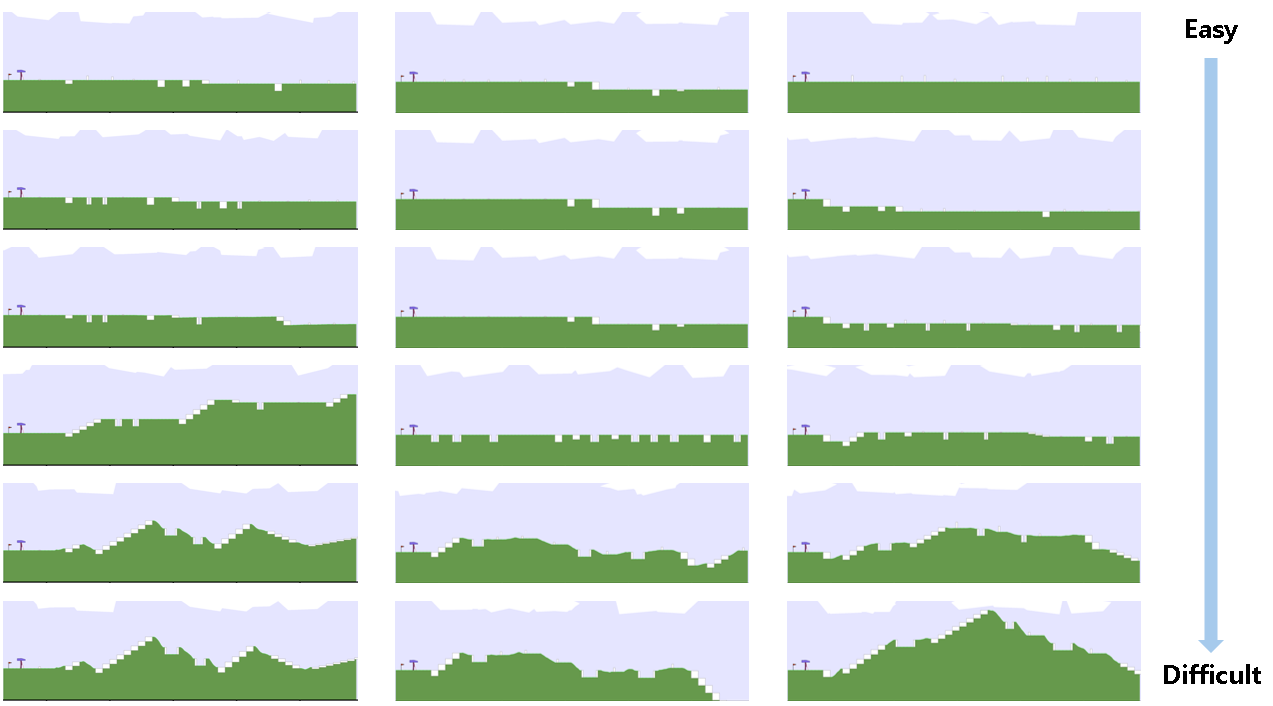}
\centering
\caption{\textbf{Controllable generation results in the 2D bipedal locomotion task}. The figure shows how the environments change as the desired difficulty level increases. We note that each column is generated using the same initial noise $\theta_T$ and random seed.}
\label{bipedal_controllable}
\end{figure}

\textbf{Controllable generation.} As done in the partially observable navigation task, we provide the results of manipulating the difficulty of the generated environments in Figure \ref{bipedal_controllable}. The results show that when we guide the generator to produce low-difficulty environments, nearly flat terrains are produced. As the desired difficulty increases, the environments become progressively more complex and challenging to walk over. This additional capability of ADD allows the trained environment critic and diffusion-based generator to be reused in applications such as benchmark generation.

\newpage
\subsection{Ablation Study}
We conduct ablation studies to analyze the role of entropy regularization term and the number of environments used in diffusion pre-training. Frist, to show whether adding an entropy term to the original UED objective plays a critical role, we measure the zero-shot generalization performance of the trained agent with varying $\omega$, which is defined in the soft UED objective $\mathop{\mathbb{E}}_{\theta \sim \Lambda}\left[\regret(\pi, \theta)\right] + \frac{1}{\omega}H(\Lambda)$. The experimental results are shown in Table \ref{omega_ablation}. From the results, we observed that performance decreases as $\omega$ becomes large. Since the influence of the entropy term diminishes as $\omega$ increases, it can be seen that our experimental results highlight the importance of the entropy term.

Next, to analyze the influence of the number of samples used during the pre-training phase, we trained the diffusion model using one million samples, which is 100 times fewer than in the original experiment, and measured the performance of the proposed algorithm. The result is a mean success rate of 0.76 $\pm\,$ 0.07 in the partially observable navigation task. This is about 11\% lower than the result reported in the original experiment. The result demonstrates that a larger number of samples used in pre-training would lead to better performance, which is quite trivial since the diffusion model can generate more diverse environments when trained with a larger number of samples. Additionally, we note that since we are dealing with an unsupervised setting and the samples used in pre-training are generated through random sampling, there is no need to worry about data scarcity.

\begin{table}[hbt!]
\caption{\textbf{Ablation study on the entropy regularization term.} The table shows the zero-shot generalization performance in the partially observable navigation task in accordance to the entropy coefficient $\omega$. We measure the average success rate over five independent seeds.}
\label{omega_ablation}
\resizebox{\textwidth}{!}{%
\begin{tabular}{p{2.5cm}|cccccc}
\hline
$\omega$     & 5                  & 10              & 20       & 40               & 80   \\ \hline
Mean success rate   & 0.85  $\pm\,$ 0.05  & 0.81  $\pm\,$ 0.05  & 0.82 $\pm\,$ 0.03  & 0.64  $\pm\,$ 0.07  & 0.47 $\pm\,$ 0.16              \\ \hline
\end{tabular}%
}
\end{table}

\newpage
\pagebreak
\section*{NeurIPS Paper Checklist}

\begin{enumerate}

\item {\bf Claims}
    \item[] Question: Do the main claims made in the abstract and introduction accurately reflect the paper's contributions and scope?
    \item[] Answer: \answerYes{} 
    \item[] Justification: The paper's contributions and scope are accurately described in the abstract and instruction.
    \item[] Guidelines:
    \begin{itemize}
        \item The answer NA means that the abstract and introduction do not include the claims made in the paper.
        \item The abstract and/or introduction should clearly state the claims made, including the contributions made in the paper and important assumptions and limitations. A No or NA answer to this question will not be perceived well by the reviewers. 
        \item The claims made should match theoretical and experimental results, and reflect how much the results can be expected to generalize to other settings. 
        \item It is fine to include aspirational goals as motivation as long as it is clear that these goals are not attained by the paper. 
    \end{itemize}

\item {\bf Limitations}
    \item[] Question: Does the paper discuss the limitations of the work performed by the authors?
    \item[] Answer: \answerYes{} 
    \item[] Justification: The limitations of the work are discussed in Section \ref{limitation}.
    \item[] Guidelines:
    \begin{itemize}
        \item The answer NA means that the paper has no limitation while the answer No means that the paper has limitations, but those are not discussed in the paper. 
        \item The authors are encouraged to create a separate "Limitations" section in their paper.
        \item The paper should point out any strong assumptions and how robust the results are to violations of these assumptions (e.g., independence assumptions, noiseless settings, model well-specification, asymptotic approximations only holding locally). The authors should reflect on how these assumptions might be violated in practice and what the implications would be.
        \item The authors should reflect on the scope of the claims made, e.g., if the approach was only tested on a few datasets or with a few runs. In general, empirical results often depend on implicit assumptions, which should be articulated.
        \item The authors should reflect on the factors that influence the performance of the approach. For example, a facial recognition algorithm may perform poorly when image resolution is low or images are taken in low lighting. Or a speech-to-text system might not be used reliably to provide closed captions for online lectures because it fails to handle technical jargon.
        \item The authors should discuss the computational efficiency of the proposed algorithms and how they scale with dataset size.
        \item If applicable, the authors should discuss possible limitations of their approach to address problems of privacy and fairness.
        \item While the authors might fear that complete honesty about limitations might be used by reviewers as grounds for rejection, a worse outcome might be that reviewers discover limitations that aren't acknowledged in the paper. The authors should use their best judgment and recognize that individual actions in favor of transparency play an important role in developing norms that preserve the integrity of the community. Reviewers will be specifically instructed to not penalize honesty concerning limitations.
    \end{itemize}

\item {\bf Theory Assumptions and Proofs}
    \item[] Question: For each theoretical result, does the paper provide the full set of assumptions and a complete (and correct) proof?
    \item[] Answer: \answerYes{} 
    \item[] Justification: We provide the full set of assumptions in Proposition \ref{soft_ued_saddlepoint}, and the complete proof is presented in Appendix \ref{appendix:saddlepoint} 
    \item[] Guidelines:
    \begin{itemize}
        \item The answer NA means that the paper does not include theoretical results. 
        \item All the theorems, formulas, and proofs in the paper should be numbered and cross-referenced.
        \item All assumptions should be clearly stated or referenced in the statement of any theorems.
        \item The proofs can either appear in the main paper or the supplemental material, but if they appear in the supplemental material, the authors are encouraged to provide a short proof sketch to provide intuition. 
        \item Inversely, any informal proof provided in the core of the paper should be complemented by formal proofs provided in appendix or supplemental material.
        \item Theorems and Lemmas that the proof relies upon should be properly referenced. 
    \end{itemize}

    \item {\bf Experimental Result Reproducibility}
    \item[] Question: Does the paper fully disclose all the information needed to reproduce the main experimental results of the paper to the extent that it affects the main claims and/or conclusions of the paper (regardless of whether the code and data are provided or not)?
    \item[] Answer: \answerYes{} 
    \item[] Justification: We provide experiment details and hyperparameters in Appendix \ref{appendix:experiments}.
    \item[] Guidelines:
    \begin{itemize}
        \item The answer NA means that the paper does not include experiments.
        \item If the paper includes experiments, a No answer to this question will not be perceived well by the reviewers: Making the paper reproducible is important, regardless of whether the code and data are provided or not.
        \item If the contribution is a dataset and/or model, the authors should describe the steps taken to make their results reproducible or verifiable. 
        \item Depending on the contribution, reproducibility can be accomplished in various ways. For example, if the contribution is a novel architecture, describing the architecture fully might suffice, or if the contribution is a specific model and empirical evaluation, it may be necessary to either make it possible for others to replicate the model with the same dataset, or provide access to the model. In general. releasing code and data is often one good way to accomplish this, but reproducibility can also be provided via detailed instructions for how to replicate the results, access to a hosted model (e.g., in the case of a large language model), releasing of a model checkpoint, or other means that are appropriate to the research performed.
        \item While NeurIPS does not require releasing code, the conference does require all submissions to provide some reasonable avenue for reproducibility, which may depend on the nature of the contribution. For example
        \begin{enumerate}
            \item If the contribution is primarily a new algorithm, the paper should make it clear how to reproduce that algorithm.
            \item If the contribution is primarily a new model architecture, the paper should describe the architecture clearly and fully.
            \item If the contribution is a new model (e.g., a large language model), then there should either be a way to access this model for reproducing the results or a way to reproduce the model (e.g., with an open-source dataset or instructions for how to construct the dataset).
            \item We recognize that reproducibility may be tricky in some cases, in which case authors are welcome to describe the particular way they provide for reproducibility. In the case of closed-source models, it may be that access to the model is limited in some way (e.g., to registered users), but it should be possible for other researchers to have some path to reproducing or verifying the results.
        \end{enumerate}
    \end{itemize}

\item {\bf Open access to data and code}
    \item[] Question: Does the paper provide open access to the data and code, with sufficient instructions to faithfully reproduce the main experimental results, as described in supplemental material?
    \item[] Answer: \answerYes{} 
    \item[] Justification: We submit the code with a detailed instruction, and all experiments are reproducible. 
    \item[] Guidelines:
    \begin{itemize}
        \item The answer NA means that paper does not include experiments requiring code.
        \item Please see the NeurIPS code and data submission guidelines (\url{https://nips.cc/public/guides/CodeSubmissionPolicy}) for more details.
        \item While we encourage the release of code and data, we understand that this might not be possible, so “No” is an acceptable answer. Papers cannot be rejected simply for not including code, unless this is central to the contribution (e.g., for a new open-source benchmark).
        \item The instructions should contain the exact command and environment needed to run to reproduce the results. See the NeurIPS code and data submission guidelines (\url{https://nips.cc/public/guides/CodeSubmissionPolicy}) for more details.
        \item The authors should provide instructions on data access and preparation, including how to access the raw data, preprocessed data, intermediate data, and generated data, etc.
        \item The authors should provide scripts to reproduce all experimental results for the new proposed method and baselines. If only a subset of experiments are reproducible, they should state which ones are omitted from the script and why.
        \item At submission time, to preserve anonymity, the authors should release anonymized versions (if applicable).
        \item Providing as much information as possible in supplemental material (appended to the paper) is recommended, but including URLs to data and code is permitted.
    \end{itemize}

\item {\bf Experimental Setting/Details}
    \item[] Question: Does the paper specify all the training and test details (e.g., data splits, hyperparameters, how they were chosen, type of optimizer, etc.) necessary to understand the results?
    \item[] Answer: \answerYes{} 
    \item[] Justification: All the training and test details are provided in Appendix \ref{appendix:experiments} and Appendix \ref{appendix:results}.
    \item[] Guidelines:
    \begin{itemize}
        \item The answer NA means that the paper does not include experiments.
        \item The experimental setting should be presented in the core of the paper to a level of detail that is necessary to appreciate the results and make sense of them.
        \item The full details can be provided either with the code, in appendix, or as supplemental material.
    \end{itemize}

\item {\bf Experiment Statistical Significance}
    \item[] Question: Does the paper report error bars suitably and correctly defined or other appropriate information about the statistical significance of the experiments?
    \item[] Answer: \answerYes{} 
    \item[] Justification: In Figure \ref{maze_result} and Figure \ref{bipedal_result}, all the results are accompanied by error bars.
    \item[] Guidelines:
    \begin{itemize}
        \item The answer NA means that the paper does not include experiments.
        \item The authors should answer "Yes" if the results are accompanied by error bars, confidence intervals, or statistical significance tests, at least for the experiments that support the main claims of the paper.
        \item The factors of variability that the error bars are capturing should be clearly stated (for example, train/test split, initialization, random drawing of some parameter, or overall run with given experimental conditions).
        \item The method for calculating the error bars should be explained (closed form formula, call to a library function, bootstrap, etc.)
        \item The assumptions made should be given (e.g., Normally distributed errors).
        \item It should be clear whether the error bar is the standard deviation or the standard error of the mean.
        \item It is OK to report 1-sigma error bars, but one should state it. The authors should preferably report a 2-sigma error bar than state that they have a 96\% CI, if the hypothesis of Normality of errors is not verified.
        \item For asymmetric distributions, the authors should be careful not to show in tables or figures symmetric error bars that would yield results that are out of range (e.g. negative error rates).
        \item If error bars are reported in tables or plots, The authors should explain in the text how they were calculated and reference the corresponding figures or tables in the text.
    \end{itemize}

\item {\bf Experiments Compute Resources}
    \item[] Question: For each experiment, does the paper provide sufficient information on the computer resources (type of compute workers, memory, time of execution) needed to reproduce the experiments?
    \item[] Answer: \answerYes{} 
    \item[] Justification: We report the type of compute workers and computation time in Appendix \ref{appendix:experiments}.
    \item[] Guidelines:
    \begin{itemize}
        \item The answer NA means that the paper does not include experiments.
        \item The paper should indicate the type of compute workers CPU or GPU, internal cluster, or cloud provider, including relevant memory and storage.
        \item The paper should provide the amount of compute required for each of the individual experimental runs as well as estimate the total compute. 
        \item The paper should disclose whether the full research project required more compute than the experiments reported in the paper (e.g., preliminary or failed experiments that didn't make it into the paper). 
    \end{itemize}
    
\item {\bf Code Of Ethics}
    \item[] Question: Does the research conducted in the paper conform, in every respect, with the NeurIPS Code of Ethics \url{https://neurips.cc/public/EthicsGuidelines}?
    \item[] Answer: \answerYes{} 
    \item[] Justification: Our research fully conform with the NeurIPS Code of Ethics.
    \item[] Guidelines:
    \begin{itemize}
        \item The answer NA means that the authors have not reviewed the NeurIPS Code of Ethics.
        \item If the authors answer No, they should explain the special circumstances that require a deviation from the Code of Ethics.
        \item The authors should make sure to preserve anonymity (e.g., if there is a special consideration due to laws or regulations in their jurisdiction).
    \end{itemize}

\item {\bf Broader Impacts}
    \item[] Question: Does the paper discuss both potential positive societal impacts and negative societal impacts of the work performed?
    \item[] Answer: \answerNA{} 
    \item[] Justification: Our work is not expected to cause direct societal impacts.
    \item[] Guidelines:
    \begin{itemize}
        \item The answer NA means that there is no societal impact of the work performed.
        \item If the authors answer NA or No, they should explain why their work has no societal impact or why the paper does not address societal impact.
        \item Examples of negative societal impacts include potential malicious or unintended uses (e.g., disinformation, generating fake profiles, surveillance), fairness considerations (e.g., deployment of technologies that could make decisions that unfairly impact specific groups), privacy considerations, and security considerations.
        \item The conference expects that many papers will be foundational research and not tied to particular applications, let alone deployments. However, if there is a direct path to any negative applications, the authors should point it out. For example, it is legitimate to point out that an improvement in the quality of generative models could be used to generate deepfakes for disinformation. On the other hand, it is not needed to point out that a generic algorithm for optimizing neural networks could enable people to train models that generate Deepfakes faster.
        \item The authors should consider possible harms that could arise when the technology is being used as intended and functioning correctly, harms that could arise when the technology is being used as intended but gives incorrect results, and harms following from (intentional or unintentional) misuse of the technology.
        \item If there are negative societal impacts, the authors could also discuss possible mitigation strategies (e.g., gated release of models, providing defenses in addition to attacks, mechanisms for monitoring misuse, mechanisms to monitor how a system learns from feedback over time, improving the efficiency and accessibility of ML).
    \end{itemize}
    
\item {\bf Safeguards}
    \item[] Question: Does the paper describe safeguards that have been put in place for responsible release of data or models that have a high risk for misuse (e.g., pretrained language models, image generators, or scraped datasets)?
    \item[] Answer: \answerNA{} 
    \item[] Justification: Our work does not have a negative use case, so we don't need special safeguards. 
    \item[] Guidelines:
    \begin{itemize}
        \item The answer NA means that the paper poses no such risks.
        \item Released models that have a high risk for misuse or dual-use should be released with necessary safeguards to allow for controlled use of the model, for example by requiring that users adhere to usage guidelines or restrictions to access the model or implementing safety filters. 
        \item Datasets that have been scraped from the Internet could pose safety risks. The authors should describe how they avoided releasing unsafe images.
        \item We recognize that providing effective safeguards is challenging, and many papers do not require this, but we encourage authors to take this into account and make a best faith effort.
    \end{itemize}

\item {\bf Licenses for existing assets}
    \item[] Question: Are the creators or original owners of assets (e.g., code, data, models), used in the paper, properly credited and are the license and terms of use explicitly mentioned and properly respected?
    \item[] Answer: \answerYes{} 
    \item[] Justification: We cite all the papers that produced the code package, and provide URLs in Appendix \ref{appendix:hyperparameters}.
    \item[] Guidelines:
    \begin{itemize}
        \item The answer NA means that the paper does not use existing assets.
        \item The authors should cite the original paper that produced the code package or dataset.
        \item The authors should state which version of the asset is used and, if possible, include a URL.
        \item The name of the license (e.g., CC-BY 4.0) should be included for each asset.
        \item For scraped data from a particular source (e.g., website), the copyright and terms of service of that source should be provided.
        \item If assets are released, the license, copyright information, and terms of use in the package should be provided. For popular datasets, \url{paperswithcode.com/datasets} has curated licenses for some datasets. Their licensing guide can help determine the license of a dataset.
        \item For existing datasets that are re-packaged, both the original license and the license of the derived asset (if it has changed) should be provided.
        \item If this information is not available online, the authors are encouraged to reach out to the asset's creators.
    \end{itemize}

\item {\bf New Assets}
    \item[] Question: Are new assets introduced in the paper well documented and is the documentation provided alongside the assets?
    \item[] Answer: \answerYes{} 
    \item[] Justification: We provide the details about training in Appendix \ref{appendix:experiments}.
    \item[] Guidelines:
    \begin{itemize}
        \item The answer NA means that the paper does not release new assets.
        \item Researchers should communicate the details of the dataset/code/model as part of their submissions via structured templates. This includes details about training, license, limitations, etc. 
        \item The paper should discuss whether and how consent was obtained from people whose asset is used.
        \item At submission time, remember to anonymize your assets (if applicable). You can either create an anonymized URL or include an anonymized zip file.
    \end{itemize}

\item {\bf Crowdsourcing and Research with Human Subjects}
    \item[] Question: For crowdsourcing experiments and research with human subjects, does the paper include the full text of instructions given to participants and screenshots, if applicable, as well as details about compensation (if any)? 
    \item[] Answer: \answerNA{} 
    \item[] Justification: Our work does not involve crowdsourcing nor research with human subjects.
    \item[] Guidelines:
    \begin{itemize}
        \item The answer NA means that the paper does not involve crowdsourcing nor research with human subjects.
        \item Including this information in the supplemental material is fine, but if the main contribution of the paper involves human subjects, then as much detail as possible should be included in the main paper. 
        \item According to the NeurIPS Code of Ethics, workers involved in data collection, curation, or other labor should be paid at least the minimum wage in the country of the data collector. 
    \end{itemize}

\item {\bf Institutional Review Board (IRB) Approvals or Equivalent for Research with Human Subjects}
    \item[] Question: Does the paper describe potential risks incurred by study participants, whether such risks were disclosed to the subjects, and whether Institutional Review Board (IRB) approvals (or an equivalent approval/review based on the requirements of your country or institution) were obtained?
    \item[] Answer: \answerNA{} 
    \item[] Justification: Our work does not involve crowdsourcing nor research with human subjects.
    \item[] Guidelines:
    \begin{itemize}
        \item The answer NA means that the paper does not involve crowdsourcing nor research with human subjects.
        \item Depending on the country in which research is conducted, IRB approval (or equivalent) may be required for any human subjects research. If you obtained IRB approval, you should clearly state this in the paper. 
        \item We recognize that the procedures for this may vary significantly between institutions and locations, and we expect authors to adhere to the NeurIPS Code of Ethics and the guidelines for their institution. 
        \item For initial submissions, do not include any information that would break anonymity (if applicable), such as the institution conducting the review.
    \end{itemize}

\end{enumerate}

\end{document}